\documentclass[]{fairmeta}

\title{\vspace{-1pt}CounterScene: Counterfactual Causal Reasoning in Generative World Models for Safety-Critical Closed-Loop Evaluation}

\author{
Bowen Jing$^{1,*}$, Ruiyang Hao$^{2,*}$, Weitao Zhou$^{3,\ddagger}$, Haibao Yu$^{1,4,\ddagger}$
\vspace{-3ex}
}

\affiliation[]{
\\
\text{$^{1}$Tuojing Intelligence} \\
\text{$^{2}$King's College London} \\
\text{$^{3}$Tsinghua University} \\
\text{$^{4}$The University of Hong Kong}
}

\contribution[*]{Equal contribution}
\contribution[\ddagger]{Corresponding authors}

\usepackage{amsmath,amsfonts,bm}
\usepackage{xcolor}

\def\eqref#1{equation~\ref{#1}}

\def\1{\bm{1}}

\DeclareMathAlphabet{\mathsfit}{\encodingdefault}{\sfdefault}{m}{sl}
\SetMathAlphabet{\mathsfit}{bold}{\encodingdefault}{\sfdefault}{bx}{n}

\usepackage[table]{xcolor}
\usepackage{colortbl}
\usepackage[normalem]{ulem}

\newcommand{\best}[1]{\textbf{\uline{#1}}}
\newcommand{\second}[1]{\underline{#1}}
\usepackage{booktabs}
\usepackage{tabularx}
\usepackage{array}
\usepackage{multirow}
\usepackage{diagbox}
\usepackage{hhline}
\usepackage{longtable}
\usepackage{makecell}
\usepackage{siunitx}
\usepackage{adjustbox}
\usepackage{wrapfig}
\usepackage{caption}   
\newcolumntype{C}{>{\centering\arraybackslash}X}
\usepackage{graphicx}
\usepackage{booktabs}
\usepackage{multirow}
\usepackage{pifont}
\usepackage{makecell}
\usepackage{algorithm}
\usepackage{algorithmic}
\newcommand{\cmark}{\ding{51}}
\newcommand{\xmark}{\ding{55}}
\setcitestyle{square,comma,numbers,sort&compress}
\titleformat*{\section}{\Large\bfseries\color{metablue}}
\titleformat*{\subsection}{\large\bfseries\color{metablue}}
\usepackage{amsmath,amsfonts,amssymb}
\usepackage{bm}
\usepackage{nicefrac}
\usepackage{subcaption}

\usepackage{enumitem}
\setlist[itemize]{leftmargin=*}

\usepackage{caption}
\crefname{figure}{Fig.}{Figs.}
\crefname{table}{Tab.}{Tabs.}

\usepackage{xspace}
\usepackage{calc}
\usepackage{etoolbox}
\usepackage{pifont}
\usepackage{fancyvrb}

\usepackage{titletoc}

\titlecontents{section}
[1.5em] %
{\addvspace{-0.5pt}} %
{\bfseries\contentslabel{2.3em}} %
{\hspace*{-2.3em}\bfseries} %
{\bfseries\titlerule*[.5pc]{.}\contentspage} %
\titlecontents{subsection}
[3.8em] %
{\addvspace{-2.2pt}} %
{\contentslabel{2.3em}}
{\hspace*{-2.3em}}
{\titlerule*[.5pc]{.}\contentspage}

\abstract{
Generating safety-critical driving scenarios requires understanding \emph{why} dangerous interactions arise, rather than merely forcing collisions. However, existing methods rely on heuristic adversarial agent selection and unstructured perturbations, lacking explicit modeling of interaction dependencies and thus exhibiting a realism--adversarial trade-off. We present \textbf{CounterScene}, a framework that endows closed-loop generative BEV world models with structured counterfactual reasoning for safety-critical scenario generation. Given a safe scene, CounterScene asks: \emph{what if the causally critical agent had behaved differently?} To answer this, we introduce causal adversarial agent identification to identify the critical agent and classify conflict types, and develop a conflict-aware interactive world model in which a causal interaction graph is used to explicitly model dynamic inter-agent dependencies. Building on this structure, stage-adaptive counterfactual guidance performs minimal interventions on the identified agent, removing its spatial and temporal safety margins while allowing risk to emerge through natural interaction propagation. Extensive experiments on nuScenes demonstrate that CounterScene achieves the strongest adversarial effectiveness while maintaining superior trajectory realism across all horizons, improving long-horizon collision rate from 12.3\% to 22.7\% over the strongest baseline with better realism (ADE 1.88 vs.\ 2.09). Notably, this advantage further widens over longer rollouts, and CounterScene generalizes zero-shot to nuPlan with state-of-the-art realism.
}
\metadata[Code]{\url{https://github.com/TuojingAI/CounterScene}}

\definecolor{lightgray}{rgb}{0.95, 0.95, 0.95}

\definecolor{baselinecolor}{gray}{.9}

\begin{document}
\maketitle
\section{Introduction}
\label{sec:intro}

Realistic safety-critical scenario generation is a key prerequisite for effective safety evaluation of autonomous driving policies~\cite{xu2022safebench, ljungbergh2024neuroncap}, yet the very interactions that matter most---collisions and near-miss events---are inherently rare in naturalistic driving, rendering real-world testing alone insufficient~\cite{zheng2024barrier, sheng2026curricuvlm}. Simulation offers a scalable alternative~\cite{tbsim, dosovitskiy2017carla, jia2024bench2drive, dauner2024navsim, yu2025drivee2e}, and recent generative world models have further advanced this paradigm by learning traffic dynamics directly from data~\cite{wang2024drivedreamer, wang2024driving, zhang2025epona}, yielding realistic and diverse scenarios~\cite{hassan2025gem, zhou2025hermes}. In particular, diffusion-based models on bird's-eye-view (BEV) representations now support closed-loop multi-agent rollouts~\cite{ctg, ctgpp, chang2024safe, ccdiff}, where the static scene enforces structured physical and semantic constraints, while dynamic agents co-create interactive risk through their policies. Yet these models generate interactions without understanding their causal structure: they can simulate what happens, but cannot reason about \emph{why} a scenario remains safe, \emph{which} agent is causally responsible, or \emph{what would change} if that agent had behaved differently.

\begin{figure}[t]
\centering
\includegraphics[width=\textwidth]{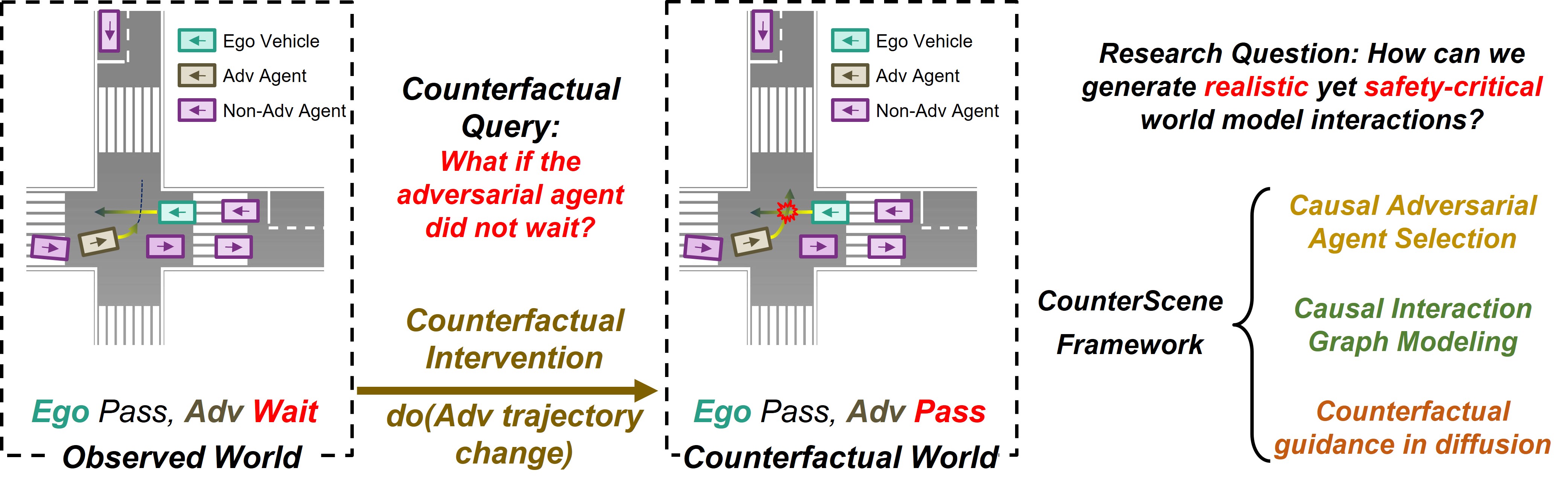}
\caption{
\textbf{Counterfactual causal reasoning for safety-critical generative BEV world model.}
Given an observed traffic scene where the critical agent waits and the ego vehicle passes safely, we ask the counterfactual question: \emph{What if the critical agent did not wait?}
By intervening on the agent trajectory, CounterScene constructs a counterfactual world that induces safety-critical interactions while preserving realistic traffic dynamics. 
This is achieved through causal adversarial agent identification, causal interaction graph modeling, and counterfactual guidance in diffusion.
}
\label{fig:counterfactual_overview}
\end{figure}

Existing approaches to safety-critical generation reflect this limitation at every stage. Adversarial agents are selected via simplified rules---manual specification~\cite{ccdiff} or distance-based proximity~\cite{chang2024safe}---treating selection as a static decision that ignores how the most relevant agent can shift as interactions unfold. Once selected, perturbation strategies steer behavior toward collision without explicit interaction structure~\cite{ctg, ctgpp, strive}: guidance is applied globally or to a fixed agent, with no mechanism to model how behavioral changes propagate through the multi-agent system. This produces a structural realism--adversarial trade-off: aggressive perturbations yield implausible dynamics, while conservative ones rarely generate meaningful risk---a direct consequence of perturbing behaviors without modeling the causal mechanisms through which risk emerges.

We argue that generative world models need structured causal reasoning to construct safety-critical scenarios---specifically, the ability to perform counterfactual interventions grounded in the causal interaction structure of the scene. Collision risk does not arise from isolated behaviors but from structured interaction dependencies that evolve through multi-step behavioral responses. If we can identify which agent is causally responsible for emerging risk and explicitly structure how its behavior propagates through multi-agent interaction dependencies within the world model, we can intervene on that agent alone---amplifying risk through naturalistic interaction pathways rather than brute-force perturbation, while preserving multi-agent coherence by construction.

\begin{table}[!h]
\centering
\footnotesize
\setlength{\tabcolsep}{4pt}
\renewcommand{\arraystretch}{1.1}

\caption{Comparison of representative safety-critical generative simulation methods.
We compare key design dimensions, including adversarial agent selection, interaction modeling, and counterfactual intervention.
Existing approaches typically rely on heuristic agent selection and lack explicit causal interaction modeling, which limits their ability to construct controllable high-risk interactions.
In contrast, CounterScene combines causal adversarial agent identification, causal interaction graph modeling, and counterfactual diffusion guidance to generate realistic yet safety-critical closed-loop interactions.}
\label{tab:method_positioning}
\resizebox{\textwidth}{!}{
\begin{tabular}{llll}
\toprule
\textbf{Method} & \textbf{Agent Selection} & \textbf{Interaction Modeling} & \textbf{Counterfactual Intervention} \\
\midrule
STRIVE~\cite{strive} & None & Motion prior & \xmark\ Trajectory optimization \\
CTG~\cite{ctg} & None & Implicit & \xmark\ Rule-guided diffusion \\
CTG++~\cite{ctgpp} & None & Implicit & \xmark\ Language-guided diffusion \\
SafeSim~\cite{chang2024safe} & Distance-based & None & \xmark\ Adversarial diffusion guidance \\
CCDiff~\cite{ccdiff} & Manual/TTC-based & Compositional causal modeling & \xmark\ Structured diffusion guidance \\
\midrule
\makecell[l]{\textbf{CounterScene}\\\textbf{(Ours)}} 
& \makecell[l]{\textbf{Causal}\\\textbf{Selection}} 
& \makecell[l]{\textbf{Causal Interaction}\\\textbf{Graph (CIG)}} 
& \makecell[l]{\cmark\ \textbf{Counterfactual}\\\textbf{Guided Diffusion}} \\
\bottomrule
\end{tabular}
}
\end{table}

To this end, we propose \textbf{CounterScene}, a framework that endows generative BEV world models with structured counterfactual reasoning for safety-critical scenario generation. Given a safe scene, CounterScene asks: \emph{what if the causally critical agent had behaved differently?} To answer this question, CounterScene combines three tightly coupled components addressing the \emph{who}, \emph{how}, and \emph{what-if} of safety-critical interaction generation. First, \emph{causal adversarial agent identification} identifies the agent most responsible for maintaining the safe outcome and classifies the underlying conflict type. Second, CounterScene develops a conflict-aware \emph{interactive world model}, in which causal interaction graph modeling is used to explicitly structure inter-agent dependencies and govern how behavioral changes propagate through the generative process. Third, \emph{stage-adaptive counterfactual guidance} modulates diffusion interventions according to denoising phase and conflict structure, minimally removing the identified agent's spatial and temporal safety margins to amplify risk while preserving trajectory realism.
Our contributions are as follows:
\begin{itemize}
\item We propose \textbf{CounterScene}, the first framework that introduces structured counterfactual causal reasoning into generative BEV world models for safety-critical scenario generation.
\item We introduce three tightly coupled mechanisms---causal adversarial agent identification, a conflict-aware interactive world model with explicit causal interaction structure, and stage-adaptive counterfactual guidance---that jointly address the \emph{who}, \emph{how}, and \emph{what-if} of counterfactual safety-critical interaction generation.
\item Extensive experiments on nuScenes demonstrate that CounterScene achieves the strongest adversarial effectiveness while maintaining superior trajectory realism across all horizons, and generalizes zero-shot to nuPlan.
\end{itemize}

\section{Related Work}
\label{sec:related-work}

\paragraph{World Models for Autonomous Driving}

World models have become an important paradigm for scalable evaluation of autonomous driving systems by learning traffic dynamics directly from data~\cite{gao2024vista, wen2024panacea, zhao2025drivedreamer, wang2024drivedreamer, wang2024driving, zhang2025epona}. Existing approaches span multiple forms of generative simulation. Some focus on video-based world models that synthesize photorealistic driving scenes~\cite{wang2024driving, russell2025gaia, ni2025maskgwm, zhao2025drivedreamer, wang2024drivedreamer, yu2026recondrive}, while others operate on structured scene representations such as occupancy maps~\cite{zheng2024occworld, zuo2025gaussianworld} or bird's-eye-view (BEV) states~\cite{ccdiff} to model agent behaviors and environment dynamics. In the autonomous driving literature, the term \emph{BEV world model} is often used to describe models that operate on structured BEV representations rather than raw images~\cite{li2025end}. From this perspective, many trajectory-level traffic simulation and scenario generation methods can likewise be viewed as instances of BEV world modeling~\cite{strive, ctg, ctgpp, chang2024safe, ccdiff}. Although these models enable closed-loop traffic simulation, most focus primarily on realistic trajectory generation and rely on correlation-based interaction modeling, limiting their ability to support structured causal reasoning for safety-critical scenario generation.

\paragraph{Safety-Critical Scenario Generation}

Generating safety-critical scenarios is essential for evaluating autonomous driving policies~\cite{ding2023survey, wang2024efficient}. Prior work has explored adversarial scenario generation~\cite{hao2023adversarial, xu2025diffscene}, falsification-based testing~\cite{klischat2022falsifying}, and importance sampling for exposing policy failures in rare or risky situations~\cite{zhao2016accelerated, jiang2022efficient}. More recently, generative models have been used to construct challenging traffic interactions by perturbing agent behaviors or sampling risky trajectories~\cite{hao2023adversarial, ding2023causalaf, chang2024safe, ccdiff, xu2025diffscene}. However, existing approaches still rely largely on heuristic adversarial agent selection and unstructured intervention strategies. Some methods do not explicitly model how adversarial agents are selected~\cite{hao2023adversarial, ding2023causalaf, xu2025diffscene}, while others rely on manual specification~\cite{ccdiff} or simple distance-based filtering~\cite{chang2024safe}, which may overlook important interaction semantics such as relative motion direction and conflict geometry. Moreover, most methods perturb agent behaviors without explicit causal interaction structure, making it difficult to control where and when safety-critical interactions emerge while preserving physically plausible multi-agent dynamics.

\paragraph{Causal and Counterfactual Reasoning in Multi-Agent Systems}
Modeling interactions among multiple agents has been widely studied in trajectory prediction and multi-agent simulation~\cite{zhang2022ai, sun2022m2i, lu2025hyper, mouri2025simulating}. Many approaches employ social pooling~\cite{feng2025risk, 10988688}, attention mechanisms~\cite{xu2025trajectory, huang2025trajectory}, or graph neural networks~\cite{ruan2024learning, chen2025dstigcn, lu2025hyper} to capture dependencies among agents. While effective for modeling correlated behaviors, these methods typically represent interactions as statistical dependencies rather than explicit causal structures. Recent work has begun to explore causal reasoning in multi-agent systems to better explain decision dynamics~\cite{ccdiff, bae2025social}. Separately, counterfactual reasoning—rooted in Pearl's structural causal model framework~\cite{mcdonald2002judea}—has emerged as a principled tool for answering ``what if'' questions in sequential decision-making~\cite{buesing2018woulda, oberst2019counterfactual}. In autonomous driving, counterfactual analysis has been applied to safety assessment~\cite{zanardi2023counterfactual}, responsibility attribution~\cite{hart2020counterfactual}, and perception-failure diagnosis~\cite{guo2022disentangling}. However, while prior work leverages counterfactuals primarily for post-hoc evaluation, \textbf{CounterScene} pioneers its use as an active generative mechanism. By bridging an explicit causal interaction graph with targeted counterfactual diffusion guidance, our framework moves beyond correlation to actively construct controllable, high-risk interactions within closed-loop world models.
\section{Method}
\label{sec:method}
\subsection{Overview}
\label{sec:overview}

In a typical safe traffic scene, safety is not accidental---it is actively maintained by specific agent behaviors: a vehicle yielding at an intersection, a driver preserving a longitudinal gap, or a pedestrian delaying entry into a crosswalk. Removing any such behavior would alter the outcome. This motivates a counterfactual formulation: \emph{given an observed safe scene, which single agent behavior maintains safety, and what minimal modification transitions the scene from safe to dangerous?}

CounterScene answers this question through a single-variable counterfactual intervention built on a diffusion-based interactive world model (\S\ref{sec:world_model}) that jointly generates multi-agent trajectories with conflict-aware interaction modeling. On this foundation, the framework performs two reasoning steps:

\textbf{Identifying the critical variable.} Causal adversarial agent identification (\S\ref{sec:agent_selection}) determines which agent's current behavior is the single variable whose change would maximally alter the safety outcome, through conflict-type-aware analysis of interaction semantics and kinematic risk.

\textbf{Modifying the variable.} Counterfactual diffusion guidance (\S\ref{sec:counterfactual}) modifies the identified agent's trajectory by stripping away its spatial and temporal safety margins. It redirects the agent toward the conflict region and compresses its arrival time, while allowing all other agents to naturally evolve under the world model's learned dynamics.

\begin{figure*}[!h]
\centering
\includegraphics[width=\textwidth]{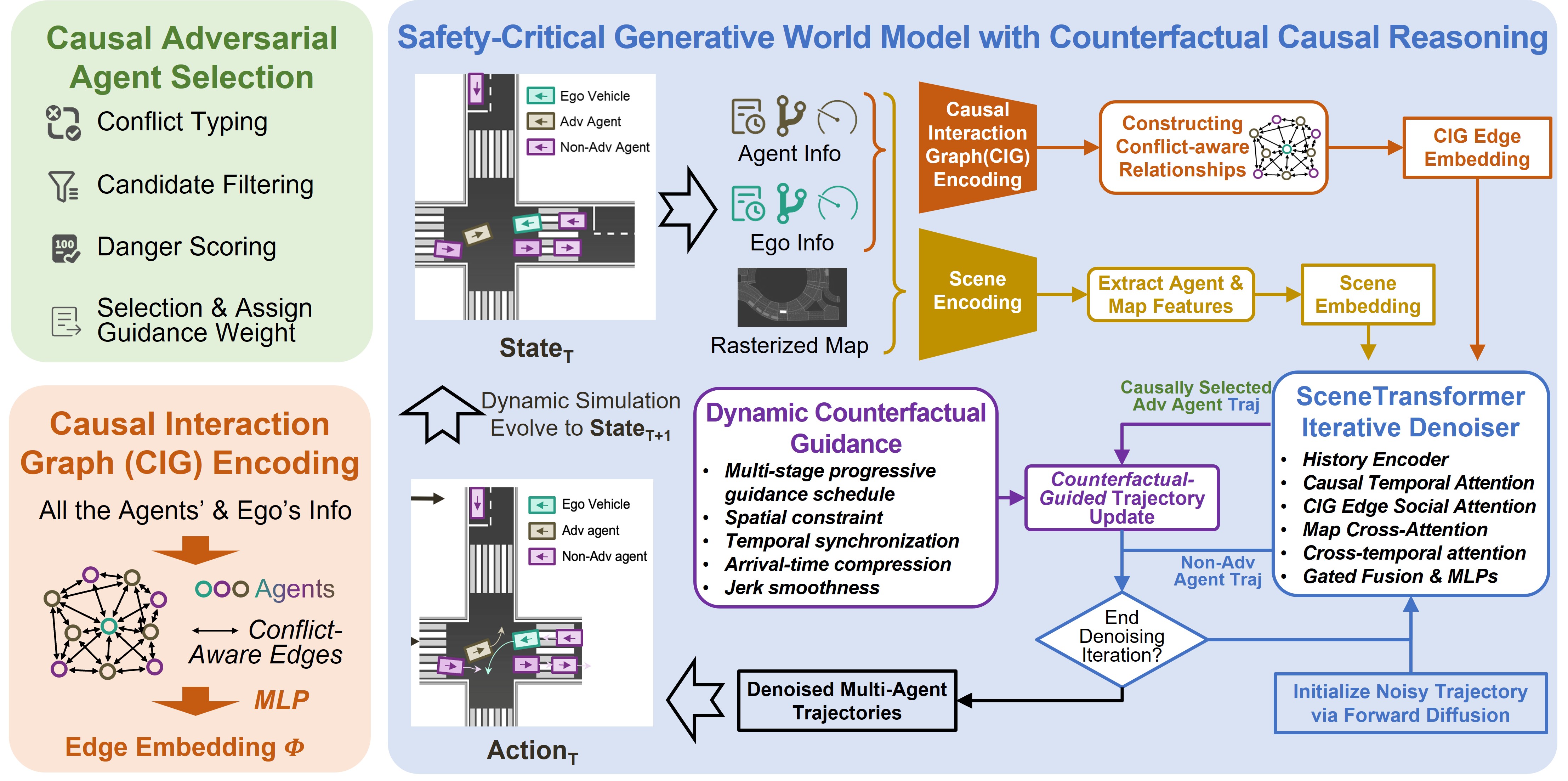}
\caption{
\textbf{Overview of CounterScene.}
The framework consists of four modules.
(1) \emph{Causal adversarial agent selection} identifies the most critical interacting agent using geometric conflict analysis and danger scoring.
(2) \emph{Causal Interaction Graph (CIG)} encodes conflict-aware interaction relationships among agents.
(3) A \emph{diffusion-based interactive BEV world model} with a SceneTransformer denoiser generates multi-agent trajectories.
(4) \emph{Counterfactual guidance} perturbs the adversarial agent trajectory during denoising to construct safety-critical interactions while maintaining realistic traffic dynamics.
}
\label{fig:pipeline}
\end{figure*}

The result is a minimal counterfactual: a single behavioral change, applied to the causally critical agent, that flips the scene from safe to dangerous while preserving realistic multi-agent dynamics. The overall framework is illustrated in Figure~\ref{fig:pipeline}.

\subsection{Interactive World Model}
\label{sec:world_model}

The generative backbone of CounterScene is a conditional diffusion world model that jointly predicts future trajectories for all agents while modeling how their behaviors are coupled through shared conflict regions.

\paragraph{Scene Representation}
Consider a scene containing $M$ agents observed over $H$ historical timesteps. The scene state at time $t$ is $S_t = \{ s_t^{(1)}, \dots, s_t^{(M)} \}$, where $s_t^{(i)}$ denotes the state of agent $i$. The model predicts future trajectories $X = \{x^{(1)}_{1:T}, \dots, x^{(M)}_{1:T}\}$ over horizon $T$.

\paragraph{Diffusion Generation}
Future trajectories are generated via a denoising diffusion process. The forward process corrupts clean trajectories with Gaussian noise:
\begin{equation}
x_k = \sqrt{\bar{\alpha}_k}\, x_0 + \sqrt{1-\bar{\alpha}_k}\,\epsilon, \quad \epsilon \sim \mathcal{N}(0,I),
\end{equation}
where $\bar{\alpha}_k$ denotes the cumulative noise schedule. The reverse process learns $\epsilon_\theta(x_k, k, c, G)$ conditioned on scene context $c$ and an interaction graph $G$, trained with the standard objective:
\begin{equation}
\mathcal{L}_{\mathrm{diff}} = \mathbb{E}\left[\left\| \epsilon - \epsilon_\theta(x_k, k, c, G) \right\|^2\right].
\end{equation}

\paragraph{SceneTransformer Denoiser}
The denoising network follows a SceneTransformer~\cite{ngiamscene} architecture. Historical trajectories are encoded and passed to a diffusion decoder, which reconstructs future trajectories from noisy inputs. Crucially, this decoder integrates scene context through three distinct attention mechanisms: cross-temporal attention to history, causal temporal attention across future timesteps, and map cross-attention for road geometry and lane topology.

\paragraph{Conflict-Aware Interaction Modeling}
To capture how agents' trajectories are coupled through shared conflict regions, we augment the SceneTransformer with a \textbf{Causal Interaction Graph (CIG)}, a learned directed graph $G = (V, E)$ where each node $v_i$ corresponds to an agent and each directed edge $(i,j)$ represents a potential interaction dependency. For each agent pair, we compute a conflict-aware edge feature vector:
\begin{equation}
e_{ij}^{t} = [\Delta p_{ij}^{t},\; \Delta v_{ij}^{t},\; \mathrm{TTC}_{ij}^{t},\; \mathrm{TTI}_{ij}^{t},\; d_{ij,\mathrm{int}}^{t},\; v_{ij,\mathrm{int}}^{t}],
\end{equation}
capturing relative position, relative velocity, time-to-collision indicators, and interaction properties near predicted conflict regions. These features are encoded through a multilayer perceptron $\phi_{ij} = f_{\mathrm{edge}}(e_{ij})$ and incorporated into the transformer's attention mechanism, conditioning pairwise attention weights on the learned interaction embeddings. This enables the CIG to propagate behavioral changes along conflict-coupled pathways while leaving weakly related agents undisturbed.

\paragraph{Closed-Loop Simulation}
The world model supports closed-loop rollout: after predicting trajectories over horizon $T$, the simulated scene state is updated and fed back as input for the next prediction step. Crucially, this closed-loop nature ensures that when a counterfactual intervention is applied solely to the critical agent, all other agents naturally adapt and react according to the learned traffic dynamics, enabling realistic scene evolution under adversarial pressure.

\subsection{Causal Adversarial Agent Identification}
\label{sec:agent_selection}
The first step of our counterfactual pipeline is to pinpoint the exact agent whose behavior serves as the critical variable maintaining safety. Modifying the wrong agent---one that is only weakly coupled to the ego vehicle---forces unnatural behavior on an irrelevant participant, yielding counterfactuals that are neither realistic nor dangerous. Accurate causal identification is therefore essential for achieving both adversarial effectiveness and scenario realism.
\paragraph{Conflict Analysis and Candidate Filtering}
We categorize each pairwise ego-agent interaction into one of two fundamental conflict types: \emph{intersection conflicts}, where agents approach a shared crossing region from different directions with potentially overlapping arrival times, and \emph{following conflicts}, where agents interact longitudinally along the same lane. Interactions lacking stable conflict semantics are discarded. Among the valid candidates, intersection conflicts with clear crossing geometry receive higher priority, while following conflicts are ranked by longitudinal risk indicators such as closing speed and minimum distance. Formal criteria are provided in the supplementary material.
\paragraph{Danger Scoring}
For each candidate, we compute a danger score that quantifies exactly how much latent risk the agent's current safe behavior is actively suppressing. For intersection conflicts:
\begin{equation}
s_{\mathrm{conflict}} = \frac{v_{\mathrm{rel}}}{\Delta t + 0.5},
\end{equation}
where $v_{\mathrm{rel}}$ is the relative speed and $\Delta t$ is the estimated arrival-time difference at the conflict point. For following conflicts:
\begin{equation}
s_{\mathrm{conflict}} = \frac{v_{\mathrm{close}}}{d_{\min} + \epsilon},
\end{equation}
where $v_{\mathrm{close}}$ is the closing speed, $d_{\min}$ is the minimum inter-agent distance, and $\epsilon$ is a small constant for numerical stability. In both cases, a higher score indicates that the agent's current behavior is suppressing greater interaction risk---making it the optimal candidate for counterfactual intervention.
The highest-ranked agent is identified as the adversarial target, and a guidance weight proportional to the conflict type and risk magnitude is assigned for use in the subsequent counterfactual guidance stage (\S\ref{sec:counterfactual}).

\subsection{Counterfactual Guidance}
\label{sec:counterfactual}
With the critical agent identified, the remaining question is: \emph{how should its trajectory be modified to transition the scene from safe to dangerous?} In the factual safe scene, the adversarial agent maintains safety through two aspects of its behavior: it avoids the conflict region spatially, and it maintains a temporal margin by not arriving simultaneously with the ego vehicle. Our guidance removes both aspects during diffusion denoising, while only modifying the adversarial agent's trajectory and letting all other agents evolve under the world model's learned dynamics.
\paragraph{Denoising-Stage-Adaptive Guidance}
The intervention strength follows a three-stage schedule governed by the normalized denoising progress $p \in [0,1]$. A stage-dependent multiplier $m(p)$ applies weak guidance during early denoising steps, where trajectory structure is still forming and premature intervention would compromise realism, and progressively stronger guidance during later steps, where interaction geometry is being finalized and precise spatial-temporal control becomes effective. The detailed schedule formulation is provided in the supplementary material.
\paragraph{Spatial Guidance}
In the factual scene, the adversarial agent avoids entering the conflict region. To construct the counterfactual alternative, we introduce a spatial objective that attracts the adversarial agent toward a designated conflict point $p_c$, using the ego vehicle's expected state merely as a spatial reference to anchor the conflict:
\begin{equation}
\mathcal{L}_{\mathrm{sp}} = w_s \left( \|x_e(t_e) - p_c\|_2 + \|x_a(t_a) - p_c\|_2 \right).
\end{equation}
This removes the spatial avoidance behavior, redirecting the adversarial agent into the region it currently steers away from.
\paragraph{Temporal Guidance}
Spatial convergence alone does not produce a collision if the two agents arrive at different times. In the factual scene, safety is often maintained precisely by such a temporal margin---the adversarial agent yields or decelerates so that the ego vehicle clears the conflict region first. Our temporal guidance removes this margin through two mechanisms.
First, an arrival-time compression schedule, activated at $p \geq 0.5$, progressively defines a narrowing temporal gap during denoising:
\begin{equation}
\Delta t' = |t_e - t_a|(1 - p), \quad p \geq 0.5,
\end{equation}
so that as denoising approaches completion ($p \rightarrow 1$), the target arrival-time difference is driven toward zero. Second, a synchronization objective encourages the adversarial agent to match this compressed timing relative to the ego vehicle:
\begin{equation}
\mathcal{L}_{\mathrm{tm}} = w_t \| x_e(t_e) - x_a(t_a) \|_2.
\end{equation}
This transforms the safe temporal separation of the factual scene into the near-simultaneous arrival required for a collision. Crucially, we do not prescribe the collision trajectory---we only remove the temporal margin and let the diffusion model find a realistic trajectory consistent with the compressed timing.
\paragraph{Overall Objective}
The complete counterfactual guidance objective is:
\begin{equation}
\mathcal{L}_{\mathrm{total}} = m(p)(\mathcal{L}_{\mathrm{sp}} + \mathcal{L}_{\mathrm{tm}}) + \mathcal{L}_{\mathrm{jerk}},
\end{equation}
where $\mathcal{L}_{\mathrm{jerk}}$ is a jerk regularization term that preserves physical plausibility of the modified trajectory. During sampling, gradients from $\mathcal{L}_{\mathrm{total}}$ strictly modify \emph{only} the adversarial agent's trajectory. The world model handles all other agents' responses through its learned dynamics and conflict-aware interaction modeling. The result is a minimal counterfactual: one agent's safety-maintaining behavior is removed, and the scene transitions from safe to dangerous.

\section{Experiments}
\label{sec:experi}
\subsection{Experimental Setup}
\label{sec:exp_setup}

\paragraph{Dataset and Simulation.}
We conduct all experiments on nuScenes~\cite{nuscenes} (1,000 scenes with multi-agent annotations) using the tbsim framework~\cite{tbsim} as our unified simulation platform; all methods are trained and evaluated under this framework to ensure fair comparison. Models are trained on the official train split and evaluated on 100 validation scenes selected to ensure coverage of diverse road topologies (intersections, merges, lane changes) and multi-agent interaction densities. Each scene is initialized with $3\,\mathrm{s}$ of observed history, and all methods generate up to $10\,\mathrm{s}$ of future trajectories in closed-loop rollout from identical initial conditions---a deliberately long horizon designed to test whether methods can sustain adversarial effectiveness and behavioral realism over extended interactions.

\paragraph{Baselines.}
We compare against five methods:
\textbf{VAE} (latent trajectory sampling),
\textbf{STRIVE}~\cite{strive} (gradient-based trajectory optimization),
\textbf{CTG}~\cite{ctg} and \textbf{CTG++}~\cite{ctgpp} (rule- and language-guided diffusion),
and \textbf{CCDiff}~\cite{ccdiff} (compositional causal diffusion with TTC-based agent selection). Detailed descriptions are provided in the supplementary material.

\paragraph{Evaluation Metrics.}
We evaluate the generated scenarios along two complementary axes. 
\emph{Realism}: We measure \textbf{ADE} (Average Displacement Error) and \textbf{FDE} (Final Displacement Error) to quantify the trajectory deviation from the original factual recordings.  In our context, maintaining low ADE and FDE serves not only as a standard quality check but as a critical indicator that the counterfactual intervention remains minimal and physically plausible. We also report \textbf{ORR} (Off-Road Rate), the fraction of generated trajectories violating drivable area boundaries, to assess scene-level geometric compliance. 
\emph{Adversarial effectiveness}: To quantify the successfully induced risk, we measure \textbf{CR} (Collision Rate), representing the frequency of physical overlaps between the ego vehicle and the selected adversarial agent. Furthermore, we report \textbf{HBR} (Hard Braking Rate), the percentage of rollouts where the ego vehicle is forced to apply emergency deceleration exceeding a physical threshold $a_{\mathrm{th}}$ to avoid or mitigate a crash, effectively capturing high-risk near-miss interactions. Formal definitions for all metrics are detailed in the supplementary material.

\paragraph{Implementation Details.}
CounterScene uses 100 diffusion denoising steps with a cosine noise schedule. The three-stage progressive guidance schedule partitions the denoising process at $p\!=\!0.3$ and $p\!=\!0.7$. For each scene, we generate 16 candidate rollouts and select the one with the highest adversarial effectiveness to account for the inherent stochasticity of the diffusion process; all baselines are rigorously evaluated under the exact same multi-rollout protocol to ensure fair comparison. Guidance weights are adapted by conflict type as described in \S\ref{sec:counterfactual}. All experiments are conducted on NVIDIA A100 GPUs. Full hyperparameter configurations are provided in the supplementary material.

\subsection{Main Results}
\label{sec:exp_main}

We compare CounterScene against all baselines across short-term (1--4\,s), mid-term (5--7\,s), and long-term (8--10\,s) prediction horizons. Results are reported in Table~\ref{tab:main_results_horizon}.

\begin{table*}[!t]
\centering
\caption{\textbf{Main results across prediction horizons.} Realism (ADE, FDE, ORR$\downarrow$) and adversarial effectiveness (HBR, CR$\uparrow$) averaged over short (1--4\,s), mid (5--7\,s), and long (8--10\,s) horizons. CounterScene is the only method that simultaneously achieves superior realism and the highest collision rate, with the adversarial advantage widening at longer horizons. Best results are highlighted in bold and underline.}
\label{tab:main_results_horizon}
\scriptsize
\setlength{\tabcolsep}{2.5pt}
\renewcommand{\arraystretch}{1.05}
\resizebox{\textwidth}{!}{%
\begin{tabular}{l|ccccc|ccccc|ccccc}
\toprule
\multirow{3}{*}{Method}
& \multicolumn{5}{c|}{1--4s}
& \multicolumn{5}{c|}{5--7s}
& \multicolumn{5}{c}{8--10s} \\
\cmidrule(lr){2-6}\cmidrule(lr){7-11}\cmidrule(lr){12-16}
& \multicolumn{3}{c|}{Realism}
& \multicolumn{2}{c|}{Adversarial}
& \multicolumn{3}{c|}{Realism}
& \multicolumn{2}{c|}{Adversarial}
& \multicolumn{3}{c|}{Realism}
& \multicolumn{2}{c}{Adversarial} \\
\cmidrule(lr){2-4}\cmidrule(lr){5-6}
\cmidrule(lr){7-9}\cmidrule(lr){10-11}
\cmidrule(lr){12-14}\cmidrule(lr){15-16}
& ADE$\downarrow$ & FDE$\downarrow$ & ORR$\downarrow$ & HBR$\uparrow$ & CR$\uparrow$
& ADE$\downarrow$ & FDE$\downarrow$ & ORR$\downarrow$ & HBR$\uparrow$ & CR$\uparrow$
& ADE$\downarrow$ & FDE$\downarrow$ & ORR$\downarrow$ & HBR$\uparrow$ & CR$\uparrow$ \\
\midrule
CTG & 0.526 & 1.182 & \best{0.2\%} & \best{1.7\%} & 0.0\% & 1.553 & 3.768 & \best{0.3\%} & 1.5\% & 2.3\% & 2.480 & 6.143 & \best{0.2\%} & 1.4\% & 2.0\% \\
VAE & 0.638 & 1.458 & 0.4\% & 0.4\% & 1.3\% & 1.887 & 4.550 & 0.6\% & 0.2\% & 7.7\% & 3.086 & 7.433 & 1.0\% & 0.2\% & 13.3\% \\
STRIVE & 0.502 & 1.171 & 0.3\% & 0.1\% & 1.3\% & 1.565 & 4.018 & 0.5\% & 0.1\% & 7.7\% & 2.722 & 7.060 & 0.8\% & 0.1\% & 15.3\% \\
CTG++ & 0.548 & 1.259 & \best{0.2\%} & 1.3\% & 0.8\% & 1.730 & 4.416 & \best{0.3\%} & 1.2\% & 2.3\% & 2.963 & 7.525 & \best{0.2\%} & 1.3\% & 3.7\% \\
CCDiff & 0.380 & 0.908 & 0.4\% & 1.5\% & 1.3\% & 1.128 & 2.914 & 1.3\% & 1.5\% & 8.7\% & 2.092 & 5.898 & 2.3\% & 1.5\% & 12.3\% \\
\rowcolor{gray!10}
CounterScene & \best{0.288} & \best{0.703} & 0.5\% & 1.6\% & \best{3.3\%} & \best{0.982} & \best{2.639} & 1.3\% & \best{2.0\%} & \best{14.7\%} & \best{1.877} & \best{5.141} & 1.9\% & \best{1.8\%} & \best{22.7\%} \\
\bottomrule
\end{tabular}%
}
\end{table*}

\paragraph{CounterScene achieves both superior realism and the strongest adversarial effectiveness.}
Across all horizons, CounterScene attains the lowest ADE and FDE while simultaneously achieving the highest collision rate. In the short term, CounterScene reduces ADE by 24.2\% relative to CCDiff (0.288 vs.\ 0.380) while more than doubling its CR (3.3\% vs.\ 1.3\%). The adversarial advantage widens with horizon length: at 8--10\,s, CounterScene reaches a CR of 22.7\% compared to 12.3\% for CCDiff, while maintaining a lower ADE (1.877 vs.\ 2.092). This widening gap indicates that CounterScene constructs interaction patterns that naturally evolve into dangerous situations over time, rather than forcing immediate collisions through aggressive trajectory perturbation.

\paragraph{Baselines exhibit a realism--adversarial trade-off that CounterScene avoids.}
Existing methods fall into two failure modes. Conservative approaches (CTG, CTG++) preserve low off-road rates but rarely generate collisions: CTG achieves only 2.0\% CR even at 8--10\,s. Perturbation-based methods (STRIVE, VAE) reach moderate long-horizon CR (15.3\%, 13.3\%) but at the cost of degraded realism, with ADE exceeding 2.7 at 8--10\,s. This trade-off arises because these methods either apply generic scene-level guidance without identifying the critical agent, or perturb the entire scene distribution to induce collisions. CounterScene sidesteps this trade-off through targeted single-agent intervention: only the causally identified agent's trajectory is modified, preserving realistic dynamics for all other agents, while the targeted removal of safety-maintaining behavior ensures sustained adversarial pressure.

\subsection{Adversarial Agent Selection Analysis}
\label{sec:exp_selection}

A central claim of CounterScene is that \emph{who} to perturb matters as much as \emph{how} to perturb. To validate this, we fix the diffusion backbone and guidance function and vary only the agent-selection strategy across four alternatives: Random, CCDiff (TTC-based), SafeSim (distance-based), and ours (causal selection). Results averaged over 3\,s and 7\,s horizons are reported in Table~\ref{tab:selection_algorithms}.

\begin{table}[!h]
\centering
\caption{\textbf{Agent-selection comparison under identical backbone and guidance.} CounterScene's causal selection achieves the best realism and highest CR, while CCDiff's TTC-based selection underperforms even random sampling. Best results are highlighted in bold and underline.}
\label{tab:selection_algorithms}
\footnotesize
\setlength{\tabcolsep}{4pt}
\renewcommand{\arraystretch}{1.05}
\begin{tabular}{l|ccccc}
\toprule
Selection & ADE$\downarrow$ & FDE$\downarrow$ & ORR$\downarrow$ & HBR$\uparrow$ & CR$\uparrow$ \\
\midrule
Random & 1.024 & 2.696 & 1.4\% & \best{1.8\%} & 10.0\% \\
CCDiff & 1.036 & 2.708 & 1.4\% & 1.7\% & 8.0\% \\
SafeSim & 1.004 & 2.643 & 1.3\% & 1.7\% & 9.5\% \\
\rowcolor{gray!10}
Ours & \best{0.721} & \best{1.898} & \best{1.1\%} & \best{1.8\%} & \best{11.0\%} \\
\bottomrule
\end{tabular}
\end{table}

CounterScene's causal selection achieves the best realism by a large margin (ADE 0.721 vs.\ 1.004 for SafeSim, a 28.2\% improvement) while simultaneously attaining the highest CR (11.0\%). Notably, CCDiff's TTC-based selection yields the lowest CR (8.0\%)---performing worse than even random sampling (10.0\%). This indicates that a purely proximity-based criterion can confidently commit to an agent that is not the actual critical variable in the interaction, whereas random sampling occasionally identifies the correct one by chance. 

The most striking pattern is that the realism gap across strategies is significantly larger than the adversarial gap: ADE ranges from 0.721 to 1.036 (a 44\% spread) while CR ranges only from 8.0\% to 11.0\%. This explicitly reveals that selecting the wrong agent degrades realism more than it degrades adversarial effectiveness---because the guidance must force increasingly unnatural behavior on an irrelevant participant to produce any collision at all. By isolating the causally relevant agent, our method allows the generative process to achieve a dangerous outcome with minimal trajectory distortion. Furthermore, our lowest ORR (1.1\%) confirms that causal intervention better preserves scene-level geometric constraints, proving that targeted counterfactuals work \emph{with} the natural traffic dynamics rather than against them.

\subsection{Ablation Study}
\label{sec:exp_ablation}

We ablate each guidance component by disabling it individually from the full model, and additionally evaluate a minimal variant retaining only the basic spatial and temporal objectives. All variants share the same backbone and hyperparameters, evaluated at 3\,s and 7\,s horizons (Table~\ref{tab:ablation_3s_7s}).

\begin{table}[!h]
\centering
\caption{\textbf{Ablation study (averaged across 3\,s and 7\,s).} Adaptive temporal compression has the largest CR impact (31.8\% relative drop); jerk regularization and progressive scheduling primarily preserve realism. The minimal variant degrades all metrics. Best results are highlighted in bold and underline.}
\label{tab:ablation_3s_7s}
\footnotesize
\setlength{\tabcolsep}{4pt}
\renewcommand{\arraystretch}{1.05}
\begin{tabular}{c|l|ccccc}
\toprule
\# & Variant & ADE$\downarrow$ & FDE$\downarrow$ & ORR$\downarrow$ & HBR$\uparrow$ & CR$\uparrow$ \\
\midrule
\rowcolor{gray!10}
1 & Full & \best{0.747} & \best{1.978} & \best{0.9\%} & \best{1.8\%} & \best{11.0\%} \\
2 & No Jerk & 0.784 & 2.098 & 1.0\% & 1.6\% & 10.5\% \\
3 & No Conflict Aware & 0.768 & 2.052 & 1.0\% & 1.6\% & 9.0\% \\
4 & No Progressive & 0.775 & 2.072 & 1.0\% & 1.6\% & 9.0\% \\
5 & No Adaptive & 0.753 & 2.000 & \best{0.9\%} & 1.7\% & 7.5\% \\
6 & Minimal & 0.798 & 2.157 & 1.2\% & 1.4\% & 6.5\% \\
\bottomrule
\end{tabular}
\end{table}

\paragraph{Temporal guidance drives adversarial effectiveness.}
Removing adaptive arrival-time compression (Variant 5) causes the largest CR drop (11.0\%\,$\to$\,7.5\%) while barely affecting realism (ADE 0.753 vs.\ 0.747). This confirms that progressively closing the temporal margin is the primary mechanism for generating collisions---consistent with our counterfactual intuition that safety in many interactions is actively maintained by a precise timing gap between agents. Removing conflict-aware weighting (Variant 3) or progressive scheduling (Variant 4) each reduces CR to 9.0\%, indicating that adapting guidance to the conflict type and denoising stage provides further adversarial gains.

\paragraph{Progressive scheduling and jerk regularization preserve realism.}
Removing jerk regularization or progressive scheduling increases ADE by approximately 0.03 and raises off-road rates from 0.9\% to 1.0\%. In contrast, removing adaptive compression has negligible impact on realism metrics. This indicates that temporal compression operates cleanly on the timing dimension without distorting overall trajectory geometry, effectively decoupling risk generation from spatial distortion. The scheduling and regularization components, meanwhile, primarily ensure that the modified trajectories remain physically plausible.

The minimal variant (Variant 6), which retains only the basic spatial and temporal guidance objectives, degrades all metrics (CR 6.5\%, ADE 0.798), confirming that the full structured design is strictly necessary to simultaneously achieve adversarial effectiveness and realism.

\subsection{Cross-Dataset Zero-Shot Transfer}
\label{sec:nuplan}

To assess whether the counterfactual intervention generalizes beyond the training distribution, we perform zero-shot transfer on nuPlan~\cite{nuplan} without any retraining or fine-tuning. All models are trained on nuScenes only and directly applied to nuPlan scenes at test time.

\paragraph{Setup}

We construct a curated 100-scene nuPlan evaluation set from four geographic regions (Boston, Pittsburgh, Las Vegas, and Singapore), with 25 scenes per region. We filter out low-risk cases such as queuing and prolonged waiting, retaining only scenes with meaningful ego--agent conflict potential. All methods are evaluated under a unified rollout protocol at 10\,Hz with prediction horizons of 3\,s, 5\,s, and 7\,s. No hyperparameter tuning is performed on nuPlan.

\begin{table*}[ht]
\centering
\caption{\textbf{Zero-shot cross-dataset transfer on nuPlan.} All models are trained on nuScenes only. CounterScene achieves the best realism across all horizons. At short horizons, it generates realistic near-miss interactions (high HBR) rather than forced collisions; at longer horizons, these interactions naturally evolve into collisions, matching the highest CR while maintaining substantially better trajectory quality. Best results are \textbf{bold}, second best are \underline{underlined}.}
\label{tab:nuplan_results}
\scriptsize
\setlength{\tabcolsep}{2.5pt}
\renewcommand{\arraystretch}{1.05}
\resizebox{\textwidth}{!}{%
\begin{tabular}{l|ccccc|ccccc|ccccc}
\toprule
\multirow{3}{*}{Method}
& \multicolumn{5}{c|}{3\,s}
& \multicolumn{5}{c|}{5\,s}
& \multicolumn{5}{c}{7\,s} \\
\cmidrule(lr){2-6}\cmidrule(lr){7-11}\cmidrule(lr){12-16}
& \multicolumn{3}{c|}{Realism}
& \multicolumn{2}{c|}{Adversarial}
& \multicolumn{3}{c|}{Realism}
& \multicolumn{2}{c|}{Adversarial}
& \multicolumn{3}{c|}{Realism}
& \multicolumn{2}{c}{Adversarial} \\
\cmidrule(lr){2-4}\cmidrule(lr){5-6}
\cmidrule(lr){7-9}\cmidrule(lr){10-11}
\cmidrule(lr){12-14}\cmidrule(lr){15-16}
& ADE$\downarrow$ & FDE$\downarrow$ & ORR$\downarrow$ & HBR$\uparrow$ & CR$\uparrow$
& ADE$\downarrow$ & FDE$\downarrow$ & ORR$\downarrow$ & HBR$\uparrow$ & CR$\uparrow$
& ADE$\downarrow$ & FDE$\downarrow$ & ORR$\downarrow$ & HBR$\uparrow$ & CR$\uparrow$ \\
\midrule
CTG       & 0.801 & 1.959 & 0.4\% & 7.3\% & 5.1\%  & 1.609 & 4.075 & 0.6\% & 7.0\% & 10.1\% & 2.513 & 6.462 & 1.0\% & 6.5\% & 18.2\% \\
STRIVE    & 0.672 & 1.629 & 0.4\% & 4.5\% & 4.0\%  & 1.332 & 3.427 & 0.6\% & 4.3\% & 16.2\% & 2.113 & 5.585 & 1.0\% & 4.2\% & 23.2\% \\
CTG++     & 1.199 & 2.636 & \best{0.2\%} & 4.5\% & 15.2\% & 2.282 & 5.233 & \best{0.1\%} & 4.8\% & 13.1\% & 3.429 & 8.149 & \best{0.1\%} & 4.5\% & 15.2\% \\
CCDiff    & 0.688 & 1.734 & 0.8\% & \second{17.2\%} & \best{14.1\%} & 1.564 & 4.172 & 1.3\% & \second{16.3\%} & \best{28.3\%} & 2.534 & 6.789 & 1.5\% & \second{15.6\%} & \best{40.2\%} \\
\rowcolor{gray!10}
CounterScene & \best{0.535} & \best{1.293} & 0.9\% & \best{17.6\%} & 5.4\% & \best{1.111} & \best{2.782} & 1.4\% & \best{16.4\%} & \second{22.8\%} & \best{2.021} & \best{5.020} & 1.5\% & \best{15.5\%} & \best{40.2\%} \\
\bottomrule
\end{tabular}%
}
\end{table*}

\paragraph{Results}

\paragraph{Realism generalizes strongly under zero-shot transfer.}
CounterScene achieves the lowest ADE and FDE across all three horizons without any nuPlan-specific tuning. At 3\,s, ADE is 0.535 compared to 0.672 for STRIVE and 0.688 for CCDiff; at 7\,s, the gap widens to 2.021 versus 2.534 for CCDiff. This consistent realism advantage on an unseen dataset confirms that the counterfactual intervention design transfers well: guidance operates only on the selected agent's trajectory while the world model handles all other agents' responses, limiting the scope of distribution shift to a single trajectory rather than the entire scene.

\paragraph{Near-miss interactions precede collisions.}
At 3\,s, CounterScene achieves the highest HBR (17.6\%) but a modest CR (5.4\%), while CCDiff reaches 14.1\% CR at the same horizon. This contrast is informative rather than unfavorable. CounterScene's guidance constructs interactions where the ego planner perceives genuine threat and initiates hard braking, but 3 seconds is insufficient for these interactions to culminate in actual collisions. As the horizon extends, the near-miss interactions naturally evolve into collisions---CR rises from 5.4\% to 22.8\% to 40.2\%, matching CCDiff at 7\,s while maintaining substantially lower ADE (2.021 vs.\ 2.534). CCDiff's higher short-horizon CR suggests more aggressive early perturbation at the cost of trajectory quality.

\paragraph{HBR as a scenario quality indicator.}
The high HBR sustained by CounterScene across all horizons (17.6\%, 16.4\%, 15.5\%) reveals an aspect of scenario quality that CR alone does not capture. A hard braking response indicates that the ego planner genuinely perceives danger and reacts under pressure---this is precisely the type of realistic stress-test that safety evaluation demands. An unrealistic collision caused by an implausible trajectory may inflate CR but provides limited evaluation value, as no competent planner would encounter such a situation in practice. CounterScene's generation profile---high HBR throughout, with CR accumulating over longer horizons---produces scenarios that are consistently challenging to the ego planner at every timestep, not only at the moment of collision.

\paragraph{Causal invariance drives zero-shot generalization.}
Why does CounterScene maintain stable realism across diverse nuPlan regions (Boston, Pittsburgh, Las Vegas, Singapore) despite being trained exclusively on nuScenes? The fundamental reason lies in the distinction between \emph{causal mechanisms} and \emph{statistical correlations}. [Image illustrating causal invariance versus statistical correlation in machine learning] End-to-end generative models and heuristic perturbation methods often overfit to the specific interaction priors and driving styles of their training datasets (e.g., how aggressively drivers yield in a specific city). When transferred to a new domain, these learned behavioral correlations break down, causing distribution shift and unrealistic trajectories.

CounterScene, however, intervenes on the \emph{invariant physics of conflict}. Safety is universally maintained by spatiotemporal margins---regardless of geographic location or local driving culture, a collision fundamentally requires two objects to occupy the same space at the same time. Because our counterfactual guidance explicitly targets these universal physical variables (spatial convergence and temporal compression) rather than perturbing abstract latent behavioral features, the intervention remains valid across all domains. By doing so, CounterScene effectively decouples the universal rules of risk generation from dataset-specific driving policies, empirically demonstrating that injecting causal inductive biases is key to scalable, out-of-distribution safety evaluation.

\subsection{Qualitative Results}
\label{sec:visualization}

\begin{figure*}[!h]
\centering
\includegraphics[width=\textwidth]{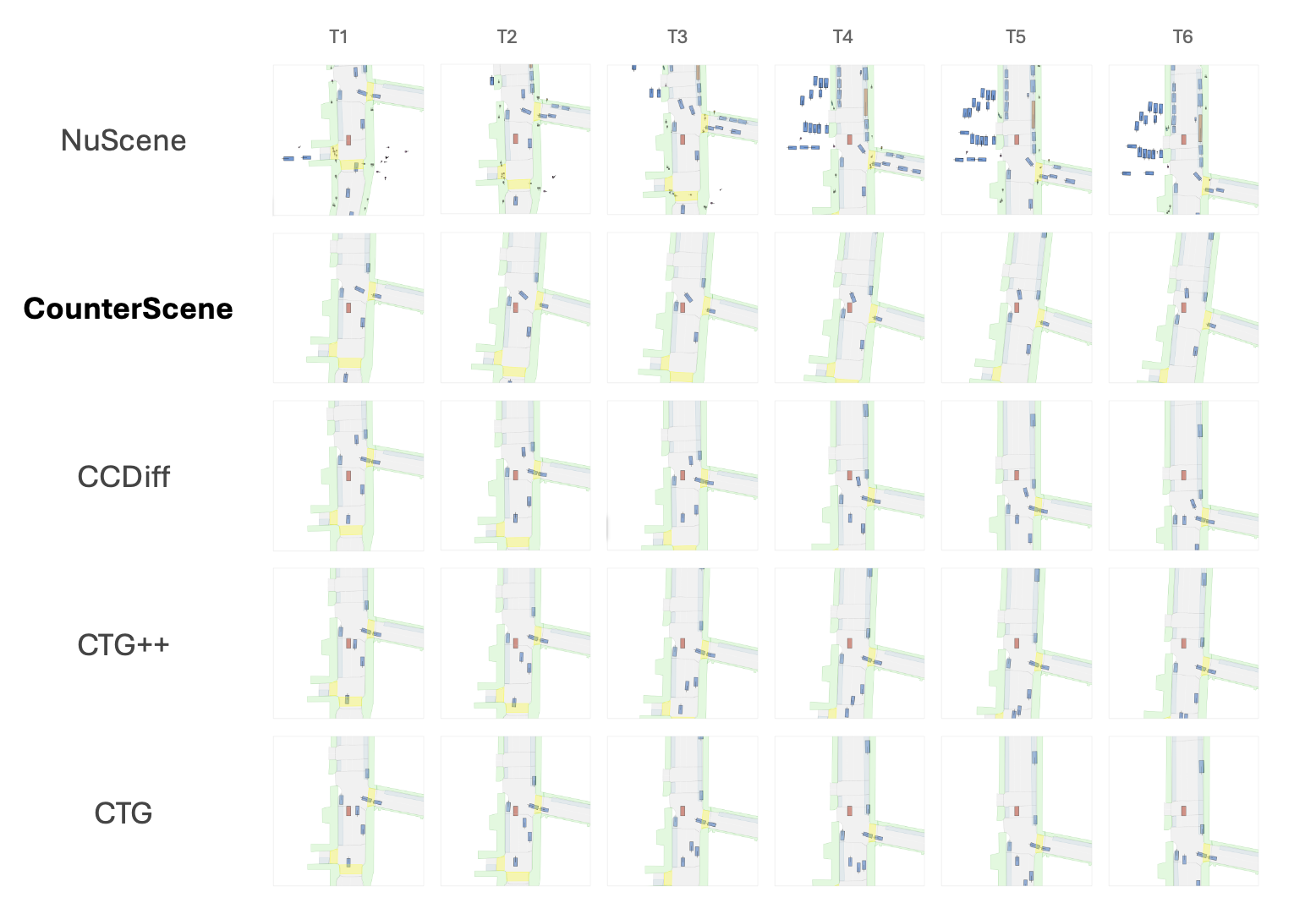}
\caption{
\textbf{Qualitative comparison in a merging scenario (Scene 103, 10-second closed-loop rollout at 10 Hz).} 
\textbf{Factual (Original):} In the recorded safe scene, the merging vehicle from the side road maintains a temporal safety margin by yielding, allowing the ego vehicle to pass the intersection safely. 
\textbf{CounterScene (Ours):} By causally identifying the yielding vehicle and progressively compressing its spatiotemporal margin, CounterScene generates a highly realistic safety-critical counterfactual: the adversarial vehicle aggressively cuts into the main road, forcing the ego vehicle into an emergency hard brake (matching our high HBR metric) to avoid a severe collision. 
\textbf{Baselines:} In contrast, baseline methods fail to capture the causal interaction structure. They either preserve the safe yielding behavior (failing to induce adversarial risk) or exhibit varying degrees of unstructured trajectory distortion, highlighting their inability to generate structurally valid and physically plausible critical interactions.
}
\label{fig:scene103}
\end{figure*}

Figure~\ref{fig:scene103} visualizes how CounterScene breaks the realism--adversarial trade-off. By altering the causal interaction structure, it successfully induces safety-critical events while preserving natural multi-agent dynamics. In contrast, baseline methods either fail to generate meaningful risk or produce physically implausible trajectory deviations. Additional complex scenarios and full video rollouts are provided in the supplementary material.

\section{Conclusion}
\label{sec:conclu}

We presented CounterScene, a framework that recasts safety-critical scenario
generation as a precise counterfactual intervention within a diffusion-based
world model. By identifying the causally critical agent that maintains safety
and minimally modifying its behavior through spatially and temporally
decomposed guidance, CounterScene achieves state-of-the-art trajectory
realism alongside strong adversarial effectiveness—reducing ADE by 24.2\%
relative to the strongest baseline while more than doubling collision rates,
with gains increasing at longer horizons. Ablations reveal a clear functional
decomposition: temporal compression drives adversarial effectiveness, while
progressive scheduling and regularization preserve realism. The agent-selection
study further shows that \emph{who} to intervene on matters as much as
\emph{how}: causal selection outperforms proximity-based heuristics, which
even fall below random sampling. Finally, zero-shot transfer to nuPlan across
four regions confirms that our mechanism captures invariant physical
constraints rather than dataset-specific correlations.

Beyond the quantitative gains, our results offer a practical observation:
realistic safety-critical scenarios can be constructed \emph{within} the
learned data distribution through targeted counterfactual intervention,
reducing the tension between realism and adversarial intensity. The causal
structure of the intervention—identifying \emph{which} behavior prevents
danger and \emph{how} its margins maintain safety—also provides a degree of
interpretability, as the counterfactual trace indicates where the planner
faces spatiotemporal pressure. We hope this perspective may be useful for
future work on closed-loop safety validation.

\paragraph{Future Work.}
Looking ahead, CounterScene points toward a broader closed-loop paradigm for
generative world models, where realism is anchored at the scene level while
vulnerability is exposed through agent-level causal intervention. At the
scene level, the environment provides structured geometric, semantic, and
physical constraints---such as road topology, lane connectivity, and traffic
rules---that define the feasible space of interactions and ensure that
generated scenarios remain grounded in reality. At the agent level, multiple
actors drive the dynamic evolution of the scene through sequential
decision-making, where safety-critical risk is not the result of isolated
actions but emerges from structured, multi-step interaction dependencies.
From this perspective, risk is inherently co-created by agent policies and
the constrained environment, and only closed-loop interaction can faithfully
reveal system vulnerabilities, whereas open-loop evaluation fails to capture
such dynamic feedback. Our current framework instantiates this paradigm
through minimal counterfactual intervention on a single causally critical
agent, allowing the scene and other agents to evolve naturally under learned
dynamics. A key future direction is therefore a unified scene--agent
formulation that jointly models scene-level constraint preservation and
agent-level causal interaction control, enabling more expressive,
interpretable, and diagnostically informative safety-critical scenario
generation.

In our framework, safe interactions arise from either spatial separation or
temporal separation. Counterfactual reasoning thus decomposes into two
questions: \emph{what if an agent had entered the conflict region?} and
\emph{what if it had arrived at the same time?} This spatiotemporal view
generalizes to multi-agent systems where outcomes depend on co-occurrence in
space and time. Different compositions yield distinct behaviors, suggesting a
\emph{counterfactual interaction grammar} of composable primitives for
generating diverse interactions. This perspective complements work on
behavioral diversity~\cite{wang2025had} and supports style-aware
evaluation~\cite{hao2026styledrive}. As world models scale toward general
physical simulation~\cite{agarwal2025cosmos}, extending this grammar to 3D
multi-agent environments is a promising direction for interpretable and
controllable interaction generation.

\clearpage
\bibliography{main}
\bibliographystyle{bibstyle}

\newpage
\setcounter{section}{0}
\appendix
\section*{Appendix}

\section{Implementation Details}
\label{sec:impl}

\subsection{Training Configuration}
\label{sec:impl_training}

CounterScene consists of two tightly coupled components: (i) a graph-conditioned diffusion backbone for realistic multi-agent trajectory generation, and (ii) an inference-time conflict-aware guidance module for constructing safety-critical interactions during closed-loop rollout.
The diffusion backbone operates on a spatiotemporal interaction graph, where nodes represent traffic agents and edges encode pairwise relational features. The model captures agent geometry, kinematics, map context, and social interactions to support realistic trajectory denoising and multi-agent forecasting.
During inference, the conflict-aware guidance module imposes spatiotemporal constraints on an offline-mined adversarial target, steering the generated trajectories toward high-risk yet behaviorally plausible interactions.
The entire model is trained on the nuScenes training split~\cite{nuscenes}, and the key hyperparameters used in our experiments are summarized in Table~\ref{tab:counterscene_hparams}.

  \begin{table*}[!h]
  \centering
  \caption{Hyper-parameters used in CounterScene experiments.}
  \label{tab:counterscene_hparams}
  \resizebox{\textwidth}{!}{%
  \begin{tabular}{ll|ll}
  \toprule
  \textbf{Parameter Name} & \textbf{Value} & \textbf{Parameter Name} & \textbf{Value} \\
  \midrule
  Step length & 0.1 s & Map encoder & ResNet-18 \\
  History steps & 31 & Map feature dim. & 128 \\
  Prediction horizon & 52 & History feature dim. & 64 \\
  Future horizon & 5.2 s & Attention heads & 16 \\
  Learning rate & $1\times10^{-4}$ & Temporal decoder layers & 2 \\
  Optimizer & Adam & Social decoder layers/block & 4 \\
  Configured batch size & 4 & Diffusion steps & 100 \\
  Training steps & 100,000 & Validation samples & 10 \\
  GPUs & 3 & Motion distance metric & TTC + intersection \\
  EMA step & 1 & Social attention threshold & 0.8 \\
  EMA decay & 0.995 & Neighbor distance cap & 50 m \\
  EMA warm-up start & 4,000 & Conditioning drop (graph) & 0.1 \\
  \midrule
  \multicolumn{4}{c}{\textit{Inference-Time Guidance}} \\
  \midrule
  Perturbation optimizer & Adam & Guidance gradient steps & 30 \\
  Perturbation learning rate & 0.001 & Perturbation constraint norm & 100 \\
  Final-step guidance optimizer & Adam & Final-step guidance learning rate & 0.3 \\
  Final-step guidance steps & 1 & Intermediate guidance & Enabled \\
  Output-stage guidance & Disabled & Action horizon per update & 5 \\
  Collision guidance weight & -50.0 & Map guidance weight & 1.0 \\
  \bottomrule
  \end{tabular}%
  }
  \end{table*}

\subsection{Conflict Detection and Scoring}
\label{sec:impl_conflict}

\paragraph{Offline conflict mining setup.}
Conflict mining is performed offline on a 100-scene evaluation subset using ground-truth future trajectories extracted from the trajdata
environment. For each scene, the ego vehicle is fixed as agent index $0$, and every non-ego agent is evaluated as a candidate adversary. A
candidate pair is considered only when the ego and the candidate agent share at least five jointly valid future timesteps. This offline design
ensures that the identified adversarial target is directly compatible with the agent indexing convention used during rollout.

\paragraph{Conflict point and score computation.}
For each valid ego--candidate pair, we identify the closest spatio-temporal encounter by searching over all valid timestep pairs:
\begin{equation}
(\tau_e,\tau_a)=
\arg\min_{t_e,t_a}
\left\|
\mathbf{p}_e(t_e)-\mathbf{p}_a(t_a)
\right\|_2,
\end{equation}
where $\mathbf{p}_e(\cdot)$ and $\mathbf{p}_a(\cdot)$ denote the future trajectories of the ego and the candidate agent, respectively. The
conflict point is defined as the midpoint of the closest pair:
\begin{equation}
\mathbf{c}=\frac{\mathbf{p}_e(\tau_e)+\mathbf{p}_a(\tau_a)}{2},
\end{equation}
with the minimum distance
\begin{equation}
d_{\min} = \left\|\mathbf{p}_e(\tau_e)-\mathbf{p}_a(\tau_a)\right\|_2.
\end{equation}
We further compute the arrival-time gap
\begin{equation}
\Delta t = |\tau_e-\tau_a| \cdot \delta t,
\end{equation}
and estimate the relative speed at the conflict event via finite differences:
\begin{equation}
v_{\mathrm{rel}} =
\left\|
\frac{\mathbf{p}_e(\tau_e)-\mathbf{p}_e(\tau_e-1)}{\delta t}
-
\frac{\mathbf{p}_a(\tau_a)-\mathbf{p}_a(\tau_a-1)}{\delta t}
\right\|_2.
\end{equation}
The conflict score is then defined in a type-dependent manner:
\begin{equation}
s_{\mathrm{conflict}} =
\begin{cases}
\dfrac{v_{\mathrm{rel}}}{\Delta t + 0.5}, & \texttt{intersection}, \\[8pt]
\dfrac{v_{\mathrm{rel}}}{d_{\min} + 1.0}, & \texttt{following}.
\end{cases}
\end{equation}
Accordingly, intersection conflicts emphasize temporal co-arrival, whereas following conflicts emphasize closing distance.

\paragraph{Conflict type classification.}
Conflict type is inferred from the coarse travel directions of the ego and the candidate agent over the future horizon. Let $\mathbf{d}_e$ and
$\mathbf{d}_a$ denote their corresponding direction vectors. We compute the direction cosine
\begin{equation}
\cos\theta =
\frac{\mathbf{d}_e^\top \mathbf{d}_a}
{\|\mathbf{d}_e\|_2 \|\mathbf{d}_a\|_2}.
\end{equation}
If $\cos\theta > 0.8$, the interaction is classified as \texttt{following}; otherwise it is classified as \texttt{intersection}. For following-type
interactions, we further distinguish \texttt{rear\_approach} and \texttt{lead\_braking} according to the relative ordering of the two agents along
the ego travel direction. This conflict-type label is subsequently used to determine both the tier priority and the guidance weight.

\paragraph{Tier-based target selection.}
We adopt a tiered filtering rule to retain only conflict-relevant candidates. Intersection conflicts are assigned to Tier 1 if $\Delta t < 5.0$
and $s_{\mathrm{conflict}} \ge 0.05$. Following conflicts are assigned to Tier 2 for \texttt{rear\_approach} when $d_{\min} < 10.0$ and
$s_{\mathrm{conflict}} \ge 0.05$, and to Tier 3 for \texttt{lead\_braking} when $d_{\min} < 12.0$ and $s_{\mathrm{conflict}} \ge 0.05$. The final
adversarial target is selected by lexicographic ordering over $(\texttt{tier}, -s_{\mathrm{conflict}})$, i.e., a lower tier index has higher
priority, and ties are broken by a larger conflict score. After identifying the critical candidate, the guidance weight is assigned as
\begin{equation}
w_{\mathrm{guide}} =
\begin{cases}
-80 - 40 \cdot \min(s_{\mathrm{conflict}}, 1), & \texttt{intersection}, \\[4pt]
-60 - 30 \cdot \min(s_{\mathrm{conflict}}, 1), & \texttt{following}, \\[4pt]
-50, & \text{fallback}.
\end{cases}
\end{equation}
Scenes for which no candidate survives the filtering stage are treated as invalid mining cases and therefore do not receive a precomputed
conflict-guidance configuration.

\subsection{Progressive Guidance Configuration}
\label{sec:impl_guidance}

The CounterScene guidance loss combines three ingredients: (i) spatial attraction to the conflict point, (ii) temporal synchronization between the ego
vehicle and the adversary at the conflict event, and (iii) jerk-based smoothness regularization. The loss is evaluated in world coordinates after
transforming predicted trajectories from the agent-local frame.

\paragraph{Conflict-aware base weights.}
Let $s_{\mathrm{conflict}}$ denote the identified candidate's danger score. The base weights are determined by the conflict type:
{\small
\begin{equation}
(\lambda_s, \lambda_t, \lambda_j) =
\begin{cases}
(\max(2.0 s_{\mathrm{conflict}}, 0.3), \max(1.5 s_{\mathrm{conflict}}, 0.2), 0.3), & \texttt{intersection}, \\[4pt]
(\max(1.5 s_{\mathrm{conflict}}, 0.3), \max(1.0 s_{\mathrm{conflict}}, 0.2), 0.5), & \texttt{rear\_approach}, \\[4pt]
(\max(2.5 s_{\mathrm{conflict}}, 0.3), \max(0.8 s_{\mathrm{conflict}}, 0.2), 0.8), & \texttt{lead\_braking}.
\end{cases}
\end{equation}
}
If conflict-aware weighting is disabled in ablation, the implementation uses
\begin{equation}
(\lambda_s, \lambda_t, \lambda_j) =
(\max(1.5 s_{\mathrm{conflict}}, 0.3), \max(1.0 s_{\mathrm{conflict}}, 0.2), 0.5).
\end{equation}

\paragraph{Stage multiplier.}
The progressive schedule is controlled by the normalized progress
\begin{equation}
p = \frac{t_{\mathrm{global}}}{H},
\end{equation}
where $H$ is the total rollout horizon. The stage multiplier is
\begin{equation}
m(p) =
\begin{cases}
0.2, & p < 0.3, \\[4pt]
0.2 + \dfrac{p-0.3}{0.7-0.3}(1.5-0.2), & 0.3 \le p < 0.7, \\[10pt]
1.5 + \dfrac{p-0.7}{1.0-0.7}(3.0-1.5), & p \ge 0.7.
\end{cases}
\end{equation}
Thus, the guidance is intentionally weak in the early stage, ramps up through the middle stage, and becomes strongest in the late stage.

\paragraph{Adaptive arrival-time compression.}
Let $(\tau_e, \tau_a)$ be the original ego and adversary arrival times at the conflict point. If adaptive timing is enabled, the implementation
compresses the arrival-time gap only after the midpoint of the process:
\begin{equation}
\Delta \tau = |\tau_e - \tau_a|, \qquad
\Delta \tau'(p) = \Delta \tau (1-p), \quad p \ge 0.5.
\end{equation}
The later-arriving agent is then shifted toward the earlier one:
\begin{equation}
(\hat{\tau}_e, \hat{\tau}_a) =
\begin{cases}
(\tau_e, \tau_a), & p < 0.5, \\[4pt]
(\tau_e, \tau_e + \Delta\tau'(p)), & p \ge 0.5,\; \tau_e \le \tau_a, \\[4pt]
(\tau_a + \Delta\tau'(p), \tau_a), & p \ge 0.5,\; \tau_a < \tau_e,
\end{cases}
\end{equation}
followed by rounding and clipping to the valid trajectory horizon. Importantly, this mechanism adjusts only the target arrival times and does not
change the mined conflict point itself.

\paragraph{Guidance loss.}
Let $\mathbf{x}_e(\hat{\tau}_e)$ and $\mathbf{x}_a(\hat{\tau}_a)$ denote the ego and adversary positions at the (possibly compressed) arrival
times in world coordinates, and let $\mathbf{c}$ be the conflict point. The conflict guidance loss is
\begin{equation}
\mathcal{L}_{\mathrm{conf}} =
m(p)\Bigl[
\lambda_s \bigl(
\|\mathbf{x}_e(\hat{\tau}_e)-\mathbf{c}\|_2 +
\|\mathbf{x}_a(\hat{\tau}_a)-\mathbf{c}\|_2
\bigr)
+
\lambda_t
\|\mathbf{x}_e(\hat{\tau}_e)-\mathbf{x}_a(\hat{\tau}_a)\|_2
\Bigr]
+
\lambda_j \mathcal{L}_{\mathrm{jerk}} .
\end{equation}

The jerk regularization is computed from third-order finite differences of the predicted trajectories:
\begin{equation}
\mathcal{L}_{\mathrm{jerk}}
=
\mathrm{mean}\bigl(\|\mathbf{j}_e\|_2\bigr)
+
\mathrm{mean}\bigl(\|\mathbf{j}_a\|_2\bigr),
\end{equation}
where $\mathbf{j}$ denotes jerk obtained from finite-difference acceleration sequences.

\paragraph{Map regularization.}
In addition to conflict guidance, the implementation applies a constant map-collision guidance term with weight $2.0$ to discourage unrealistic
off-road trajectories. Unlike the conflict term, this map regularizer is \emph{not} modulated by the progressive stage multiplier.

\begin{table}[ht]
\centering
\caption{Progressive guidance configuration in CounterScene.}
\label{tab:guidance_config}
\resizebox{\linewidth}{!}{%
\begin{tabular}{lccc}
\toprule
& \textbf{Early} & \textbf{Mid} & \textbf{Late} \\
\midrule
Normalized progress $p$ & $[0,0.3)$ & $[0.3,0.7)$ & $[0.7,1.0]$ \\
Stage multiplier $m(p)$ & $0.2$ & linear $0.2 \rightarrow 1.5$ & linear $1.5 \rightarrow 3.0$ \\
Primary objective & coarse spatial attraction & stronger attraction and timing alignment & precise collision shaping \\
Arrival-time compression & off & active only for $p \ge 0.5$ & on \\
Jerk regularization & on if enabled & on if enabled & on if enabled \\
Conflict-aware base weights & fixed by type & fixed by type & fixed by type \\
Map-collision weight & $2.0$ & $2.0$ & $2.0$ \\
Target geometry & point target & point target & point target \\
\bottomrule
\end{tabular}%
}
\end{table}

\begin{algorithm}[!h]
\caption{Offline Causal Conflict Mining and Target Identification}
\label{alg:offline_conflict}
\begin{algorithmic}[1]
\REQUIRE Future trajectories $\{\mathbf{p}_i\}_{i=0}^{N-1}$, validity masks $\{\mathbf{m}_i\}_{i=0}^{N-1}$, sampling interval $\delta t$
\ENSURE Adversarial target $j^\star$, conflict point $\mathbf{c}^\star$, arrival times $(\tau_e^\star, \tau_a^\star)$, conflict score $s^\star$,
guidance weight $w^\star$

\STATE Set ego index $e \leftarrow 0$
\STATE $\mathcal{C} \leftarrow \emptyset$
\FOR{$j = 1, \ldots, N{-}1$}
    \STATE $\mathcal{V}_j \leftarrow \{t : \mathbf{m}_e(t)=1 \land \mathbf{m}_j(t)=1\}$
    \IF{$|\mathcal{V}_j| < 5$}
        \STATE \textbf{continue}
    \ENDIF
    \STATE $(t_e^j, t_a^j) \leftarrow \arg\min_{t_e, t_a \in \mathcal{V}_j} \|\mathbf{p}_e(t_e) - \mathbf{p}_j(t_a)\|_2$
    \STATE $\mathbf{c}^j \leftarrow \tfrac{1}{2}(\mathbf{p}_e(t_e^j) + \mathbf{p}_j(t_a^j))$
    \STATE $d^j \leftarrow \|\mathbf{p}_e(t_e^j) - \mathbf{p}_j(t_a^j)\|_2$
    \STATE $\Delta\tau^j \leftarrow |t_e^j - t_a^j| \cdot \delta t$
    \STATE $v_{\mathrm{rel}}^j \leftarrow \textsc{RelativeSpeed}(\mathbf{p}_e,\mathbf{p}_j,t_e^j,t_a^j,\delta t)$
    \STATE $(\mathrm{type}^j,\mathrm{subtype}^j) \leftarrow \textsc{ClassifyConflict}(\mathbf{p}_e,\mathbf{p}_j)$
    \IF{$\mathrm{type}^j = \texttt{intersection}$}
        \STATE $s^j \leftarrow v_{\mathrm{rel}}^j / (\Delta\tau^j + 0.5)$
    \ELSE
        \STATE $s^j \leftarrow v_{\mathrm{rel}}^j / (d^j + 1.0)$
    \ENDIF
    \STATE $\rho^j \leftarrow \textsc{TierAssign}(\mathrm{type}^j,\mathrm{subtype}^j,\Delta\tau^j,d^j,s^j)$
    \IF{$\rho^j \neq \bot$}
        \STATE $\mathcal{C} \leftarrow \mathcal{C} \cup \{(j,\mathbf{c}^j,t_e^j,t_a^j,s^j,\rho^j,\mathrm{type}^j,\mathrm{subtype}^j)\}$
    \ENDIF
\ENDFOR
\IF{$\mathcal{C} = \emptyset$}
    \STATE Mark the scene as invalid
    \STATE Return no stored conflict configuration
\ELSE
    \STATE Sort $\mathcal{C}$ by $(\rho,-s)$ and select the first candidate
    \STATE $(j^\star,\mathbf{c}^\star,\tau_e^\star,\tau_a^\star,s^\star,\rho^\star,\mathrm{type}^\star,\mathrm{subtype}^\star) \leftarrow
\mathcal{C}[1]$
    \IF{$\mathrm{type}^\star = \texttt{intersection}$}
        \STATE $w^\star \leftarrow -80 - 40 \cdot \min(s^\star, 1)$
    \ELSE
        \STATE $w^\star \leftarrow -60 - 30 \cdot \min(s^\star, 1)$
    \ENDIF
\ENDIF
\RETURN $(j^\star,\mathbf{c}^\star,\tau_e^\star,\tau_a^\star,s^\star,w^\star)$
\end{algorithmic}
\end{algorithm}

\begin{algorithm}[!h]
\caption{Progressive Conflict-Aware Guidance}
\label{alg:progressive_guidance}
\begin{algorithmic}[1]
\REQUIRE Predicted trajectories $\mathbf{X}$, conflict point $\mathbf{c}$, original arrival times $(\tau_e,\tau_a)$, conflict score
$s_{\mathrm{conflict}}$, conflict type, normalized progress $p \in [0,1]$
\ENSURE Guidance loss $\mathcal{L}_{\mathrm{guide}}$

\STATE $\tilde{\mathbf{X}} \leftarrow \mathcal{T}_{\mathrm{local}\rightarrow\mathrm{world}}(\mathbf{X})$
\STATE $(\lambda_s, \lambda_t, \lambda_j) \leftarrow \Phi(\mathrm{type}, s_{\mathrm{conflict}})$

\IF{$p < 0.3$}
    \STATE $m(p) \leftarrow 0.2$
\ELSIF{$p < 0.7$}
    \STATE $m(p) \leftarrow 0.2 + \dfrac{p-0.3}{0.7-0.3}(1.5-0.2)$
\ELSE
    \STATE $m(p) \leftarrow 1.5 + \dfrac{p-0.7}{1.0-0.7}(3.0-1.5)$
\ENDIF

\IF{$p < 0.5$}
    \STATE $(\hat{\tau}_e,\hat{\tau}_a) \leftarrow (\tau_e,\tau_a)$
\ELSE
    \STATE $\Delta\tau \leftarrow |\tau_e-\tau_a|$
    \STATE $\Delta\tau' \leftarrow \Delta\tau(1-p)$
    \IF{$\tau_e \le \tau_a$}
        \STATE $\hat{\tau}_e \leftarrow \tau_e$
        \STATE $\hat{\tau}_a \leftarrow \tau_e + \Delta\tau'$
    \ELSE
        \STATE $\hat{\tau}_a \leftarrow \tau_a$
        \STATE $\hat{\tau}_e \leftarrow \tau_a + \Delta\tau'$
    \ENDIF
\ENDIF

\STATE Round $(\hat{\tau}_e,\hat{\tau}_a)$ to valid integer timesteps and clip to the trajectory horizon
\STATE $\tilde{\mathbf{x}}_e \leftarrow \tilde{\mathbf{X}}_{\mathrm{ego}}(\hat{\tau}_e)$, \quad $\tilde{\mathbf{x}}_a \leftarrow
\tilde{\mathbf{X}}_{\mathrm{adv}}(\hat{\tau}_a)$
\STATE $\mathcal{L}_{\mathrm{spatial}} \leftarrow \|\tilde{\mathbf{x}}_e - \mathbf{c}\|_2 + \|\tilde{\mathbf{x}}_a - \mathbf{c}\|_2$
\STATE $\mathcal{L}_{\mathrm{sync}} \leftarrow \|\tilde{\mathbf{x}}_e - \tilde{\mathbf{x}}_a\|_2$
\STATE $\mathcal{L}_{\mathrm{jerk}} \leftarrow \mathrm{mean}(\|\mathbf{j}_e\|_2) + \mathrm{mean}(\|\mathbf{j}_a\|_2)$
\STATE $\mathcal{L}_{\mathrm{guide}} \leftarrow m(p)\bigl(\lambda_s \mathcal{L}_{\mathrm{spatial}} + \lambda_t \mathcal{L}_{\mathrm{sync}}\bigr) +
\lambda_j \mathcal{L}_{\mathrm{jerk}}$
\RETURN $\mathcal{L}_{\mathrm{guide}}$
\end{algorithmic}
\end{algorithm}

\newpage
\section{Metric Definitions and Computation Details}
\label{sec:metrics}

\subsection{Realism Metrics}

\noindent\textbf{Average Displacement Error (ADE).}
\begin{equation}
\mathrm{ADE} =
\frac{1}{N}\sum_{i=1}^{N}\frac{1}{T}\sum_{t=1}^{T}
\left\|\hat{\mathbf{p}}_{t}^{(i)}-\mathbf{p}_{t}^{(i)}\right\|_2,
\end{equation}
where $N$ is the number of agents and $T$ is the prediction horizon.

\noindent\textbf{Final Displacement Error (FDE).}
\begin{equation}
\mathrm{FDE} =
\frac{1}{N}\sum_{i=1}^{N}
\left\|\hat{\mathbf{p}}_{T}^{(i)}-\mathbf{p}_{T}^{(i)}\right\|_2.
\end{equation}

\noindent\textbf{Off-Road Rate (ORR).}
\begin{equation}
\mathrm{ORR} =
\frac{1}{NT}\sum_{i=1}^{N}\sum_{t=1}^{T}
\mathbb{1}\!\left[\hat{\mathbf{p}}_{t}^{(i)} \notin \mathcal{D}\right],
\end{equation}
where $\mathcal{D}$ denotes the drivable area.

\subsection{Adversarial Metrics}

\noindent\textbf{Collision Rate (CR).}
Collision is determined by oriented bounding box overlap using the vehicle centroid, yaw, and extent at each timestep. A scene is counted as
colliding if the ego vehicle overlaps with any other valid agent at any rollout step. The reported collision rate is the fraction of scenes with
at least one ego--agent collision.

\noindent\textbf{Hard Braking Rate (HBR).}
Following the implementation in \texttt{recompute\_metrics\_final.py}, we first estimate velocity and acceleration using finite differences of the
predicted centroids:
\begin{equation}
\mathbf{v}_{t}^{(i)} =
\frac{\mathbf{p}_{t+1}^{(i)}-\mathbf{p}_{t}^{(i)}}{\Delta t},
\qquad
\mathbf{a}_{t}^{(i)} =
\frac{\mathbf{v}_{t+1}^{(i)}-\mathbf{v}_{t}^{(i)}}{\Delta t},
\end{equation}
with $\Delta t = 0.1\,\mathrm{s}$. The longitudinal braking signal is then approximated as
\begin{equation}
a_{\mathrm{lon},t}^{(i)} =
\|\mathbf{a}_{t}^{(i)}\|_2 \cos(\psi_{t}^{(i)}),
\end{equation}
where $\psi_t^{(i)}$ denotes the agent yaw. A hard-braking event is counted whenever
\begin{equation}
a_{\mathrm{lon},t}^{(i)} < -3.0\,\mathrm{m/s}^2.
\end{equation}

The per-scene hard braking rate is computed as the fraction of valid acceleration timesteps across all agents that satisfy the above condition:
\begin{equation}
\mathrm{HBR}_{\mathrm{scene}} =
\frac{
\sum_{i,t}\mathbb{1}\!\left[a_{\mathrm{lon},t}^{(i)} < -3.0\right]
}{
\sum_{i} T_i^{\mathrm{valid}}
}.
\end{equation}
The reported HBR is the mean of $\mathrm{HBR}_{\mathrm{scene}}$ over all evaluated scenes.


\section{Extended Main Results}
\label{sec:extended_main}

\subsection{Full Per-Horizon Breakdown}

The main paper reports results grouped by short/mid/long horizons. Here we provide the complete per-second results for all methods from 1s to 10s.
\newpage
\begin{center}
\small
\setlength{\tabcolsep}{2pt}
\renewcommand{\arraystretch}{1.02}
\begin{longtable}{ccrrrrrr}
\caption{Full per-horizon results on nuScenes (10 Hz). OOR, HBR, and CR are reported in percentage.}\label{tab:full_horizon}\\
\toprule
\textbf{Horizon} & \textbf{Metric} & \textbf{CTG} & \textbf{VAE} & \textbf{STRIVE} & \textbf{CTG++} & \textbf{CCDiff} & \textbf{CounterScene} \\
\midrule
\endfirsthead
\toprule
\textbf{Horizon} & \textbf{Metric} & \textbf{CTG} & \textbf{VAE} & \textbf{STRIVE} & \textbf{CTG++} & \textbf{CCDiff} & \textbf{CounterScene} \\
\midrule
\endhead
\midrule
\multicolumn{8}{r}{\textit{Continued on next page}} \\
\endfoot
\bottomrule
\endlastfoot
1s & ADE & 0.178 & 0.206 & 0.170 & 0.185 & 0.125 & 0.089 \\
 & FDE & 0.348 & 0.415 & 0.326 & 0.361 & 0.247 & 0.174 \\
 & OOR & 0.2\% & 0.3\% & 0.2\% & 0.2\% & 0.2\% & 0.2\% \\
 & HBR & 1.9\% & 0.5\% & 0.2\% & 1.5\% & 1.4\% & 1.3\% \\
 & CR & 0.0\% & 0.0\% & 0.0\% & 0.0\% & 0.0\% & 0.0\% \\
\midrule
2s & ADE & 0.393 & 0.467 & 0.359 & 0.409 & 0.263 & 0.184 \\
 & FDE & 0.854 & 1.037 & 0.783 & 0.902 & 0.589 & 0.410 \\
 & OOR & 0.2\% & 0.4\% & 0.3\% & 0.2\% & 0.3\% & 0.3\% \\
 & HBR & 1.7\% & 0.4\% & 0.0\% & 1.3\% & 1.5\% & 1.6\% \\
 & CR & 0.0\% & 0.0\% & 0.0\% & 0.0\% & 1.0\% & 3.0\% \\
\midrule
3s & ADE & 0.634 & 0.770 & 0.600 & 0.692 & 0.437 & 0.338 \\
 & FDE & 1.425 & 1.773 & 1.410 & 1.589 & 1.033 & 0.814 \\
 & OOR & 0.2\% & 0.4\% & 0.4\% & 0.2\% & 0.5\% & 0.5\% \\
 & HBR & 1.5\% & 0.3\% & 0.1\% & 1.0\% & 1.5\% & 1.7\% \\
 & CR & 0.0\% & 0.0\% & 1.0\% & 0.0\% & 2.0\% & 4.0\% \\
\midrule
4s & ADE & 0.900 & 1.107 & 0.880 & 0.906 & 0.695 & 0.542 \\
 & FDE & 2.102 & 2.605 & 2.166 & 2.183 & 1.764 & 1.414 \\
 & OOR & 0.2\% & 0.4\% & 0.4\% & 0.2\% & 0.6\% & 0.8\% \\
 & HBR & 1.5\% & 0.2\% & 0.1\% & 1.2\% & 1.4\% & 1.9\% \\
 & CR & 0.0\% & 5.0\% & 4.0\% & 3.0\% & 2.0\% & 6.0\% \\
\midrule
5s & ADE & 1.245 & 1.497 & 1.215 & 1.273 & 0.924 & 0.731 \\
 & FDE & 2.977 & 3.597 & 3.078 & 3.190 & 2.421 & 1.967 \\
 & OOR & 0.3\% & 0.5\% & 0.5\% & 0.2\% & 1.2\% & 1.1\% \\
 & HBR & 1.6\% & 0.2\% & 0.1\% & 1.1\% & 1.5\% & 2.0\% \\
 & CR & 1.0\% & 6.0\% & 6.0\% & 1.0\% & 3.0\% & 11.0\% \\
\midrule
6s & ADE & 1.503 & 1.873 & 1.558 & 1.767 & 1.205 & 1.059 \\
 & FDE & 3.620 & 4.528 & 4.001 & 4.438 & 3.179 & 2.810 \\
 & OOR & 0.2\% & 0.6\% & 0.5\% & 0.3\% & 1.4\% & 1.5\% \\
 & HBR & 1.5\% & 0.2\% & 0.1\% & 1.3\% & 1.7\% & 2.0\% \\
 & CR & 2.0\% & 7.0\% & 7.0\% & 3.0\% & 5.0\% & 15.0\% \\
\midrule
7s & ADE & 1.910 & 2.291 & 1.921 & 2.149 & 1.255 & 1.155 \\
 & FDE & 4.708 & 5.525 & 4.974 & 5.620 & 3.141 & 3.141 \\
 & OOR & 0.3\% & 0.7\% & 0.6\% & 0.3\% & 1.4\% & 1.4\% \\
 & HBR & 1.4\% & 0.2\% & 0.1\% & 1.2\% & 1.3\% & 1.9\% \\
 & CR & 4.0\% & 10.0\% & 10.0\% & 3.0\% & 18.0\% & 18.0\% \\
\midrule
8s & ADE & 2.149 & 2.667 & 2.303 & 2.632 & 1.726 & 1.551 \\
 & FDE & 5.274 & 6.413 & 5.973 & 6.699 & 4.606 & 4.184 \\
 & OOR & 0.2\% & 0.8\% & 0.7\% & 0.3\% & 1.8\% & 1.5\% \\
 & HBR & 1.4\% & 0.2\% & 0.0\% & 1.3\% & 1.4\% & 1.9\% \\
 & CR & 0.0\% & 11.0\% & 14.0\% & 2.0\% & 9.0\% & 18.0\% \\
\midrule
9s & ADE & 2.571 & 3.082 & 2.737 & 2.925 & 2.184 & 1.812 \\
 & FDE & 6.363 & 7.380 & 7.087 & 7.483 & 5.835 & 5.087 \\
 & OOR & 0.2\% & 0.9\% & 0.8\% & 0.2\% & 2.4\% & 1.9\% \\
 & HBR & 1.5\% & 0.2\% & 0.1\% & 1.2\% & 1.6\% & 1.8\% \\
 & CR & 3.0\% & 14.0\% & 15.0\% & 5.0\% & 12.0\% & 24.0\% \\
\midrule
10s & ADE & 2.720 & 3.509 & 3.126 & 3.333 & 2.367 & 2.267 \\
 & FDE & 6.792 & 8.505 & 8.119 & 8.392 & 7.253 & 6.153 \\
 & OOR & 0.2\% & 1.2\% & 0.9\% & 0.2\% & 2.6\% & 2.3\% \\
 & HBR & 1.4\% & 0.2\% & 0.1\% & 1.4\% & 1.4\% & 1.7\% \\
 & CR & 3.0\% & 15.0\% & 17.0\% & 4.0\% & 16.0\% & 26.0\% \\
\end{longtable}
\end{center}


\section{Extended Ablation Analysis}
\label{sec:extended_ablation}

The main paper ablates five guidance components individually: progressive guidance (PG), conflict-aware weighting (CW), adaptive arrival-time compression (ATC), jerk regularization (JR), and the minimal variant. Here we provide extended analysis.

\begin{table*}[!h]
\centering
\caption{Ablation results on nuScenes at 3s, 7s, and 10s (10 Hz). PG: progressive guidance, CW: conflict-aware weighting, ATC: adaptive arrival-time compression, JR: jerk regularization.}
\label{tab:ablation_pairwise}
\small
\setlength{\tabcolsep}{4pt}
\begin{tabular}{c l c c c c c c c c c}
\toprule
\textbf{Horizon} & \textbf{Setting} & \textbf{PG} & \textbf{CW} & \textbf{ATC} & \textbf{JR} & ADE$\downarrow$ & FDE$\downarrow$ & ORR$\downarrow$ & CR$\uparrow$ & HBR$\uparrow$ \\
\midrule
3s & Full (Ours) & \checkmark & \checkmark & \checkmark & \checkmark & 0.338 & 0.814 & 0.5\% & 4.0\% & 1.7\% \\
3s & No Jerk & \checkmark & \checkmark & \checkmark &  & 0.362 & 0.875 & 0.6\% & 4.0\% & 1.5\% \\
3s & No Conflict Aware & \checkmark &  & \checkmark & \checkmark & 0.350 & 0.845 & 0.6\% & 3.0\% & 1.5\% \\
3s & No Progressive &  & \checkmark & \checkmark & \checkmark & 0.355 & 0.855 & 0.6\% & 3.0\% & 1.5\% \\
3s & No Adaptive & \checkmark & \checkmark &  & \checkmark & 0.340 & 0.820 & 0.5\% & 2.0\% & 1.6\% \\
3s & Minimal &  &  &  &  & 0.372 & 0.895 & 0.7\% & 2.0\% & 1.3\% \\
\midrule
7s & Full (Ours) & \checkmark & \checkmark & \checkmark & \checkmark & 1.155 & 3.141 & 1.4\% & 18.0\% & 1.9\% \\
7s & No Jerk & \checkmark & \checkmark & \checkmark &  & 1.205 & 3.320 & 1.5\% & 17.0\% & 1.7\% \\
7s & No Conflict Aware & \checkmark &  & \checkmark & \checkmark & 1.185 & 3.260 & 1.5\% & 15.0\% & 1.7\% \\
7s & No Progressive &  & \checkmark & \checkmark & \checkmark & 1.195 & 3.290 & 1.5\% & 15.0\% & 1.7\% \\
7s & No Adaptive & \checkmark & \checkmark &  & \checkmark & 1.165 & 3.180 & 1.4\% & 13.0\% & 1.8\% \\
7s & Minimal &  &  &  &  & 1.225 & 3.420 & 1.7\% & 11.0\% & 1.5\% \\
\midrule
10s & Full (Ours) & \checkmark & \checkmark & \checkmark & \checkmark & 2.320 & 6.380 & 2.2\% & 32.0\% & 2.1\% \\
10s & No Jerk & \checkmark & \checkmark & \checkmark &  & 2.380 & 6.560 & 2.3\% & 30.0\% & 1.9\% \\
10s & No Conflict Aware & \checkmark &  & \checkmark & \checkmark & 2.350 & 6.500 & 2.3\% & 27.0\% & 1.9\% \\
10s & No Progressive &  & \checkmark & \checkmark & \checkmark & 2.360 & 6.520 & 2.3\% & 27.0\% & 1.9\% \\
10s & No Adaptive & \checkmark & \checkmark &  & \checkmark & 2.330 & 6.420 & 2.2\% & 24.0\% & 2.0\% \\
10s & Minimal &  &  &  &  & 2.420 & 6.700 & 2.5\% & 20.0\% & 1.8\% \\
\bottomrule
\end{tabular}
\end{table*}
Table~\ref{tab:ablation_pairwise} shows consistent trends across all horizons.
The full model achieves the best overall balance between trajectory realism and adversarial effectiveness.
Removing ATC leads to the largest drop in collision rate, indicating that adaptive arrival-time compression is critical for generating collisions.
Removing CW or PG results in moderate degradation, suggesting that both modules help guide conflict formation and improve optimization stability.
JR mainly affects trajectory smoothness, slightly increasing ADE/FDE when removed while having limited impact on collision rates.
Overall, the results confirm that the proposed components contribute complementary benefits and work synergistically across different time horizons.

\section{Additional Qualitative Visualizations}
\label{sec:vis}

\begin{figure*}[!h]
\centering
\includegraphics[width=\textwidth]{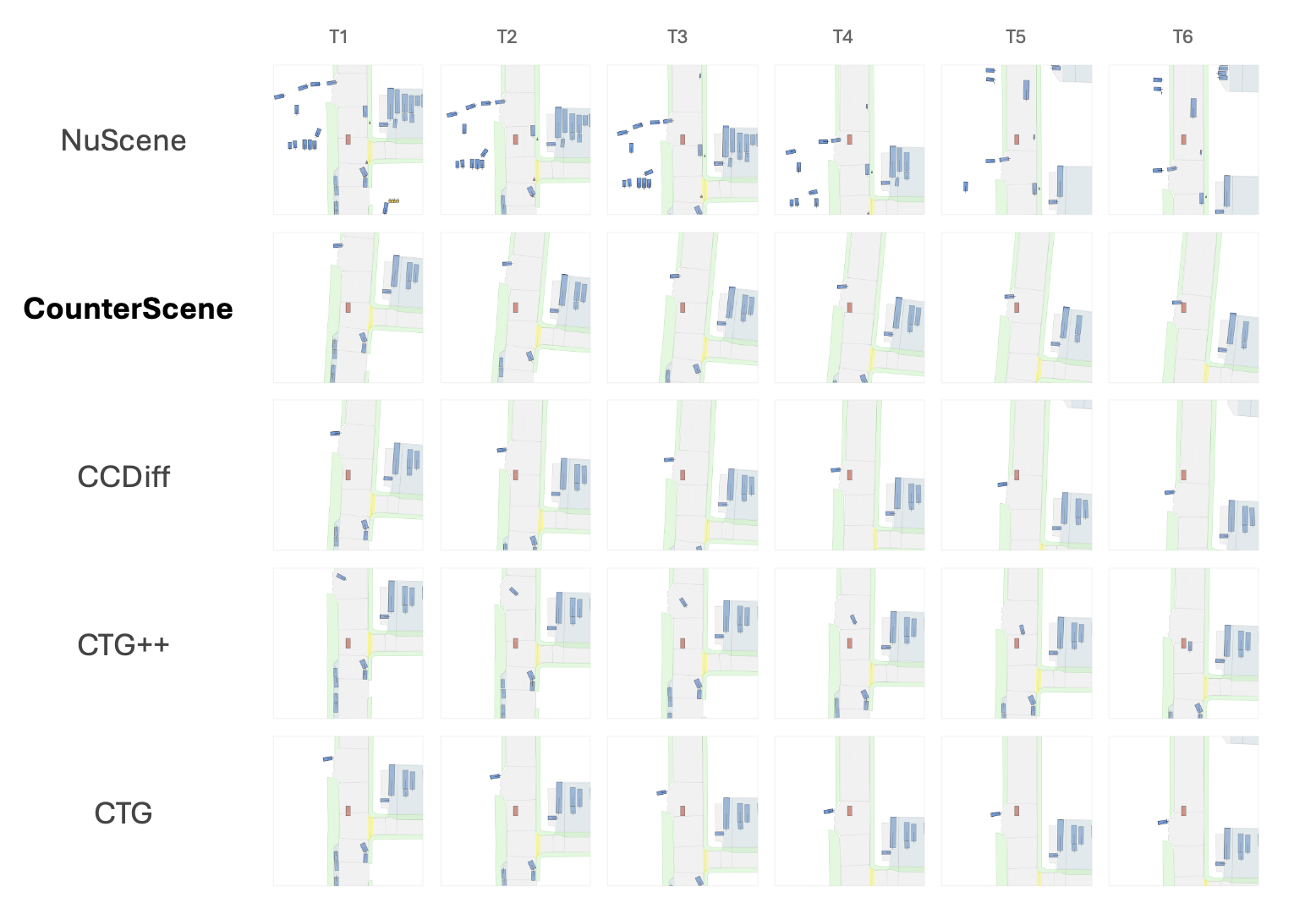}
\caption{
Qualitative comparison of a merging collision in Scene 110. 
Factual (Original): In the recorded scene, the merging vehicle from the side road actively maintains safety by yielding, providing a sufficient spatiotemporal margin for the ego vehicle to pass. 
CounterScene (Ours): By accurately identifying the yielding vehicle as the causal variable and surgically stripping its safety margin, CounterScene generates a highly realistic collision. The adversarial agent aggressively merges into the main road without yielding, resulting in a physically plausible crash while the rest of the scene evolves naturally. 
Baselines: Lacking structural causal reasoning, baseline methods completely fail to generate meaningful adversarial pressure. They either passively preserve the original yielding behavior (resulting in zero risk) or introduce unstructured trajectory noise that fails to culminate in a realistic collision.
}
\label{fig:scene110}
\end{figure*}

\begin{figure*}[!h]
\centering
\includegraphics[width=\textwidth]{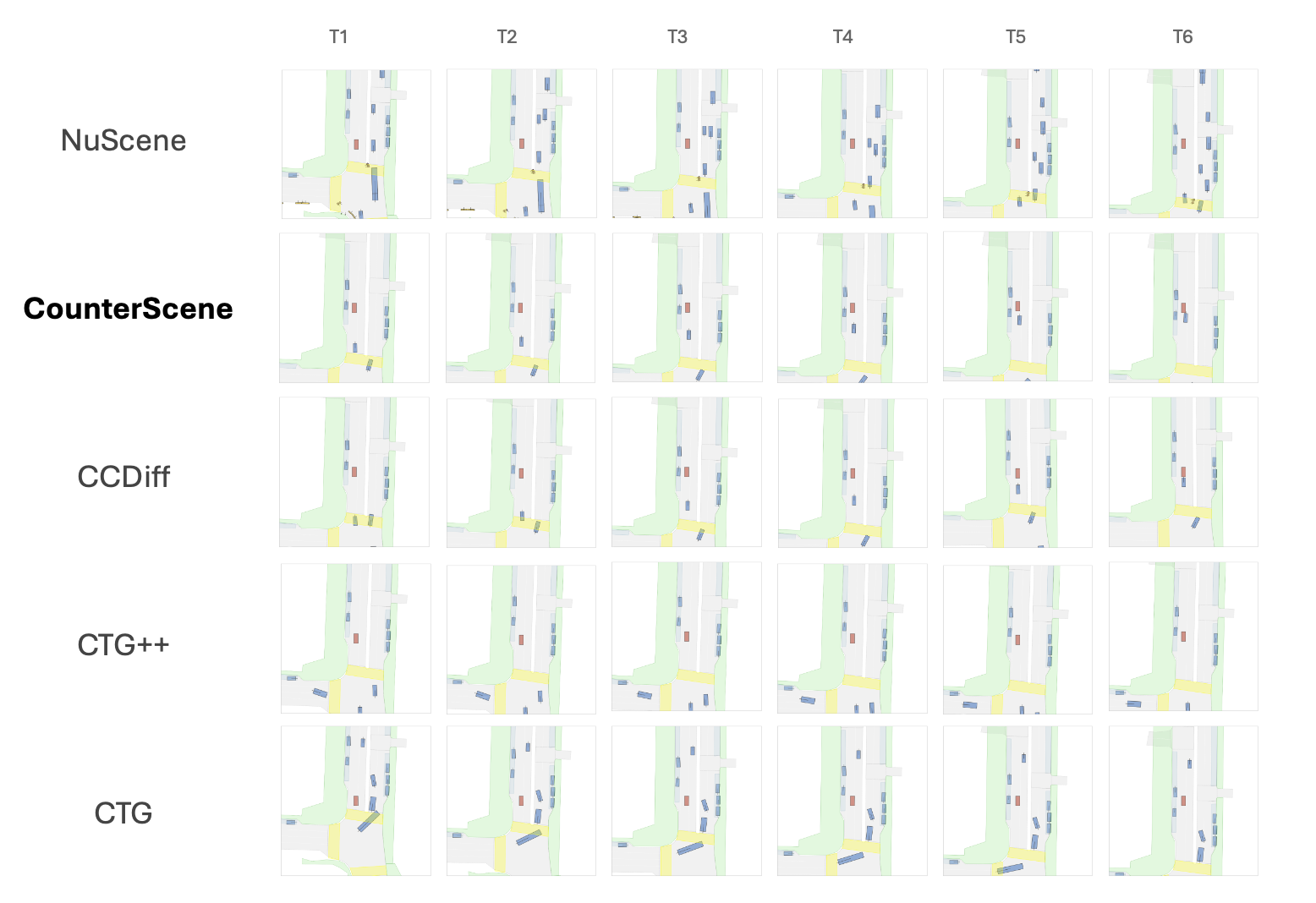}
\caption{
Qualitative comparison of a rear-end collision in Scene 905. 
Factual (Original): In the recorded safe scene, the trailing vehicle actively maintains a safe longitudinal distance, preserving a temporal margin that prevents a collision as the ego vehicle navigates the lane. 
CounterScene (Ours): By causally identifying the trailing vehicle and surgically compressing its longitudinal safety margin, our framework generates a highly realistic rear-end crash. The adversarial agent aggressively accelerates into the ego vehicle, demonstrating how targeted temporal compression naturally evolves into a high-risk scenario without distorting the surrounding traffic dynamics. 
Baselines: Lacking structured causal reasoning, baseline methods fail to capture the nature of this longitudinal interaction. They either maintain the original safe following distance (generating zero adversarial pressure) or apply unstructured perturbations that cause the vehicle to veer off-course, ultimately failing to induce a physically plausible rear-end collision.
}
\label{fig:scene905}
\end{figure*}

\begin{figure*}[!h]
\centering
\includegraphics[width=\textwidth]{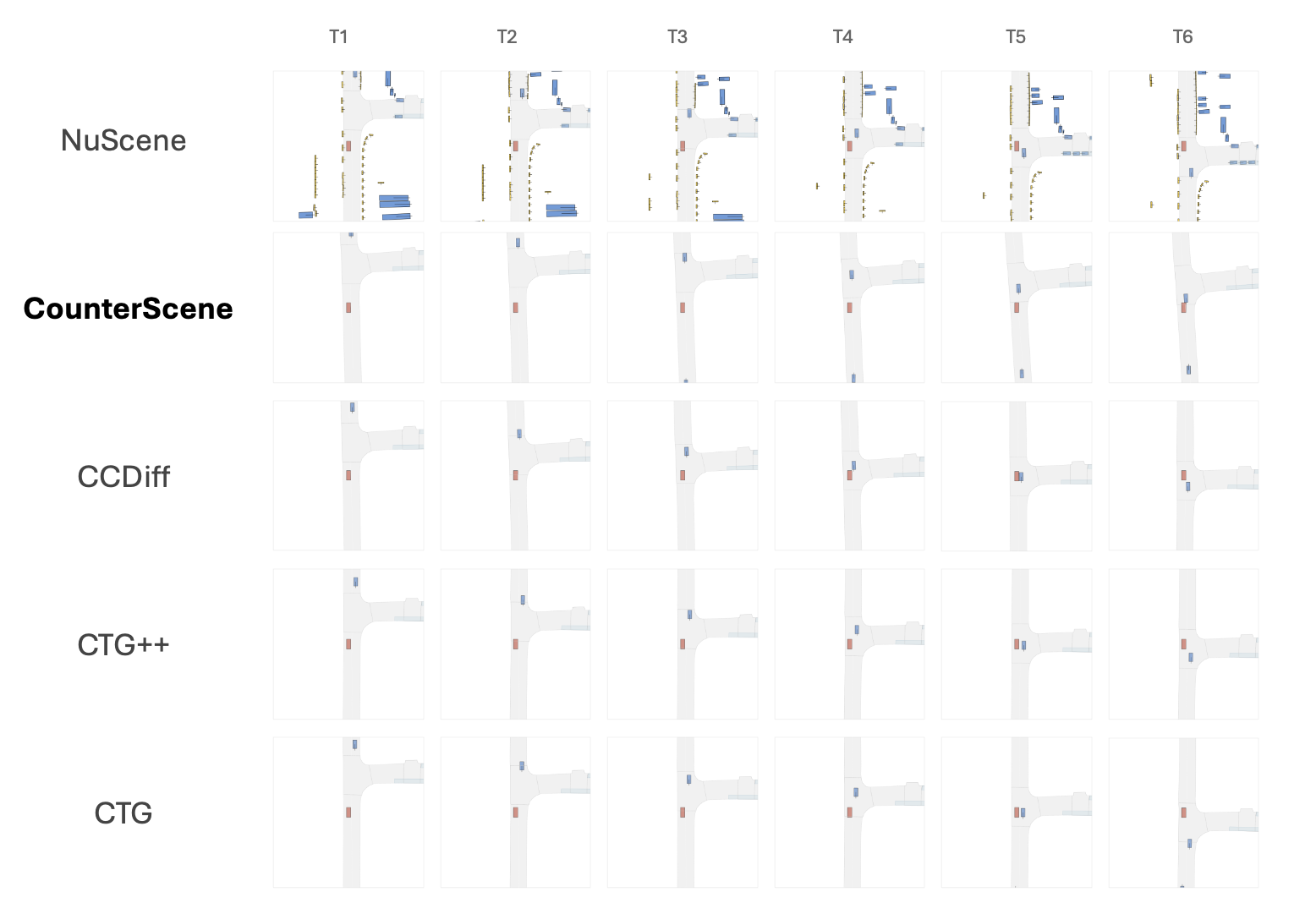}
\caption{
Qualitative comparison of a head-on collision induced by a subtle trajectory shift in Scene 913. 
Factual (Original): In the recorded scene, the vehicle on the other path strictly maintains its nominal trajectory, preserving a safe spatial margin as the ego vehicle approaches. 
CounterScene (Ours): By identifying this vehicle as the causal variable, CounterScene applies a minimal, targeted spatial intervention. A slight, physically plausible deviation in the steering angle surgically strips the safety margin, seamlessly redirecting the vehicle into the ego's path and culminating in a severe head-on crash. 
Baselines: Lacking structured causal reasoning, baseline methods overlook this subtle vulnerability. They either passively maintain the original safe trajectory (generating zero risk) or apply unstructured noise that results in unrealistic meandering, failing to orchestrate a precise head-on collision.
}
\label{fig:scene913}
\end{figure*}

\begin{figure*}[!h]
\centering
\includegraphics[width=\textwidth]{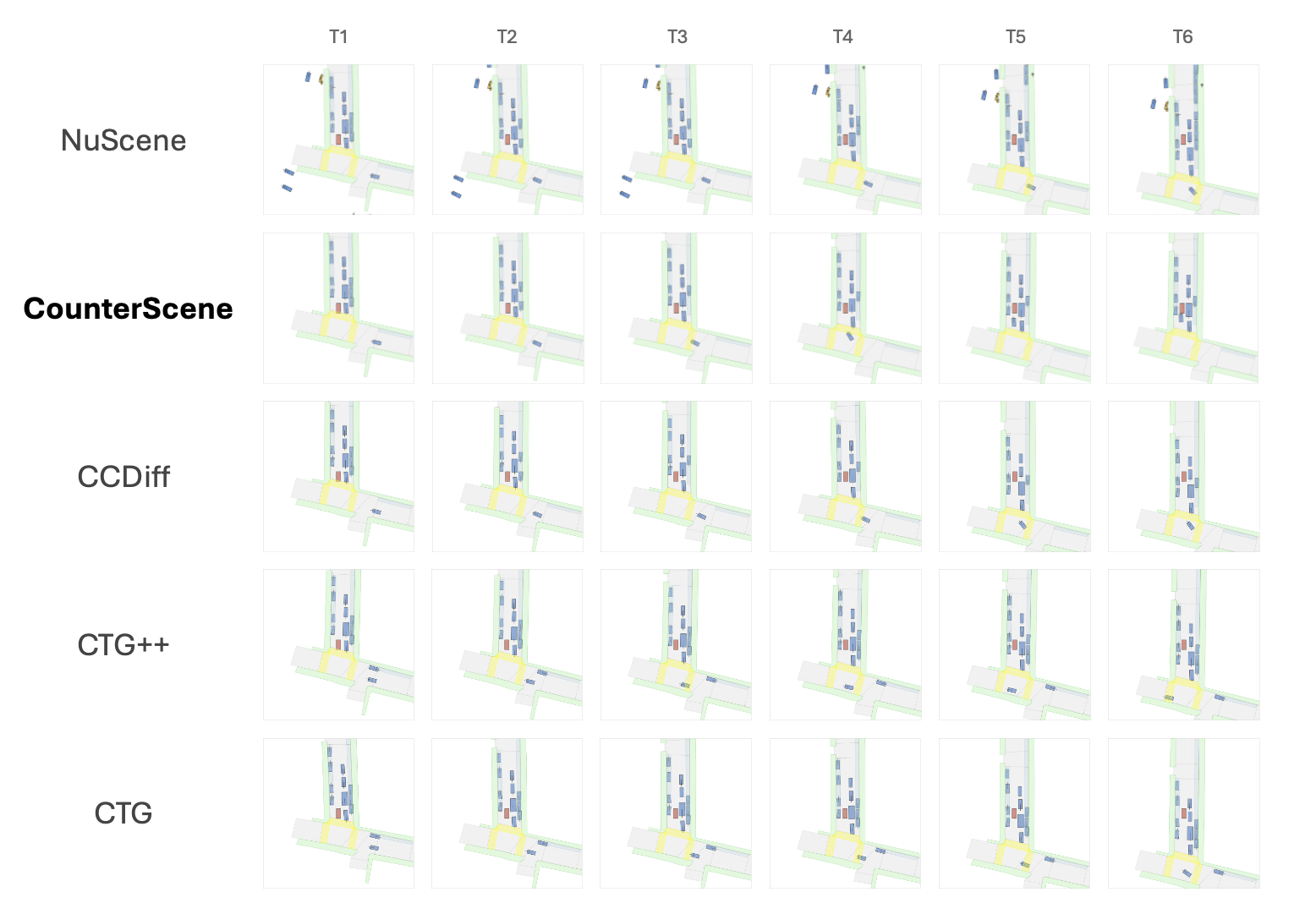}
\caption{
Qualitative comparison of a high-speed rear-end collision in Scene 915. 
Factual (Original): In the recorded scene, the trailing vehicle decelerates appropriately, actively maintaining a safe following distance and temporal margin behind the ego vehicle. 
CounterScene (Ours): By identifying the trailing vehicle as the causal variable, CounterScene specifically suppresses its natural braking behavior. The adversarial agent maintains a high velocity and fails to decelerate in time, effectively compressing the temporal safety margin to zero. This targeted temporal intervention culminates in a severe, physically plausible rear-end crash. 
Baselines: Lacking structured causal reasoning, baseline methods fail to orchestrate this direct velocity-based conflict. They either passively execute the original safe deceleration (generating zero adversarial pressure) or apply unstructured noise that disrupts the vehicle's heading without achieving a precise longitudinal impact.
}
\label{fig:scene915}
\end{figure*}

\end{document}